\documentclass{article}
\PassOptionsToPackage{numbers, compress}{natbib}

\usepackage[preprint]{neurips_2026}

\usepackage[utf8]{inputenc} 
\usepackage[T1]{fontenc}    
\usepackage{hyperref}       
\usepackage{url}            
\usepackage{booktabs}       
\usepackage{amsfonts}       
\usepackage{nicefrac}       
\usepackage{microtype}      
\usepackage{xcolor}         

\usepackage{graphicx}
\usepackage{subfigure}
\usepackage{booktabs} 

\usepackage{amsmath}
\usepackage{amssymb}
\usepackage{mathtools}
\usepackage{amsthm}
\usepackage{makecell}
\usepackage{color}
\usepackage{array}
\usepackage{caption}
\usepackage{graphicx}
\usepackage{subcaption}
\usepackage{tcolorbox}
\usepackage{multirow}
\usepackage{nccmath}
\usepackage{booktabs}
\usepackage{wrapfig}

\theoremstyle{plain}

\theoremstyle{definition}

\theoremstyle{remark}

\newtheorem*{remark*}{Remark}
\newcommand{\M}{Crys-JEPA}

\usepackage{bbm}
\usepackage{bm}
\usepackage[numbers]{natbib}

\title{\M: Accelerating Crystal Discovery via Embedding Screening and Generative Refinement}

\author{
Nian Liu$^{1}$ \quad Nikita Kazeev$^{1}$ \quad Stephen Gregory Dale$^{1}$ \quad Artem Maevskiy$^{1}$ \vspace{0.10cm}\\
\textbf{Yuwei Zeng}$^{1}$ \quad \textbf{Ryoji Kubo}$^{1}$  \quad \textbf{Pengru Huang}$^{1}$ \quad \textbf{Thomas Laurent}$^{2}$ \vspace{0.10cm}\\
\textbf{Yann LeCun}$^{3\ 4}$ \quad \textbf{Kostya S. Novoselov}$^{1}$ \quad \textbf{Xavier Bresson}$^{1}$
\vspace{0.15cm}\\
\texttt{\{nianliu, yuweizeng, ryojikubo\}@u.nus.edu, tlaurent@lmu.edu, yann@cs.nyu.edu}\vspace{0.0cm}\\
\texttt{\{kna, sdale, maevskiy, pengru, kostya, xaviercs\}@nus.edu.sg }\vspace{0.0cm}\\
{\normalfont $^1$National University of Singapore \hspace{0.1cm} $^2$Loyola Marymount University \hspace{0.1cm} $^3$New York University \hspace{0.1cm} $^4$AMI}\\
}

\begin{document}
\maketitle
\begin{abstract}
De novo crystal generation seeks to discover materials that are not merely realistic, but also stable and novel. However, most existing generative models are trained to maximize the likelihood of observed crystals, which encourages samples to stay close to known materials yet not necessarily align with the criteria that matter in discovery. 
Our empirical analysis shows that current crystal generative models exhibit a clear conflict between stability and novelty: samples near the observed distribution tend to retain stability but offer limited novelty, whereas samples farther from it often lose stability rapidly.
This suggests that the useful region for discovering crystals that are both stable and novel is extremely narrow. 
To move beyond this limitation,
we introduce \M, a joint embedding predictive architecture for crystals that learns an energy-aware latent space preserving formation-energy differences. 
In this space, stability assessment can be reformulated as an embedding-based comparison against accessible training crystals, reducing the reliance on expensive energy evaluation and task-specific external references. Building on \M, we further develop a screening-and-refinement pipeline that identifies promising generated crystals and reintroduces them to refine the generative model. On MP-20 and Alex-MP-20 datasets, we achieve improvements over baselines up to 53.8\% and 72.7\% on \texttt{V.S.U.N} metric, respectively. Our code is available at \url{https://github.com/liun-online/Crys_JEPA}.
\end{abstract}

\section{Introduction}
\label{intro}
Discovering new materials is a key driver of progress across a wide range of applications, including solar cells, batteries, and catalysis~\cite{butler2018machine}. Among material classes, crystals are of particular importance because their periodic atomic arrangements give rise to diverse and tunable physical properties, making them a central target in computational materials design. This motivates the task of \textit{de novo generation} (DNG)~\cite{cdvae}, which aims to discover entirely new crystal structures without relying on predefined templates. In recent years, DNG has been substantially advanced by deep generative models, particularly diffusion~\cite{ddpm} and flow matching~\cite{lipman2022flow}.

Most existing DNG models~\cite{cdvae, mattergen, adit} are trained under the conventional objective of maximizing the log-likelihood of the observed data. However, it remains unclear to what extent this objective improves the criteria that matter most in practice, namely \textit{Validity} (V), \textit{Stability} (S), \textit{Uniqueness} (U), and \textit{Novelty} (N), which we collectively denote as V.S.U.N. In Section~\ref{tradeoff}, we begin with an empirical study showing that current crystal generative models generally exhibit a pronounced trade-off between stability and novelty. To better understand this phenomenon, we analyze how these metrics vary with respect to the density landscape of the observed distribution. Our results suggest that crystal generation imposes an exceptionally strict precision requirement: moving closer to high-density regions is often insufficient to improve novelty, whereas drifting toward low-density regions can already destroy stability. In other words, there is little effective intermediate region in which a crystal can be both stable and novel. Consequently, although log-likelihood maximization encourages generated samples to stay near the observed distribution, these nearby regions do not necessarily satisfy both stability and novelty.

Notably, standard training datasets consist primarily of already known stable crystals. Therefore, without changing the underlying generative backbone, one natural way to alleviate the stability--novelty trade-off is to reintroduce crystals that satisfy V.S.U.N into training, especially those that are both stable and novel, so that the model can fit a more desirable distribution. When external data cannot be used due to fairness considerations, an alternative is to identify such promising crystals directly from model generations. This, however, requires reliable V.S.U.N assessment for generated samples. While validity, uniqueness, and novelty can be evaluated relatively cheaply, stability is substantially more difficult to assess. Stability is defined through a formation-energy comparison against a reference set, which introduces two fundamental challenges. First, \textit{reference ambiguity}: the choice of reference depends on the task, and even public references such as the Materials Project~\cite{MaterialsProject} evolve over time, making it impossible to know the final reference set in advance during training. Second, \textit{computational cost}: standard stability evaluation typically relies on density functional theory (DFT), which is prohibitively expensive at scale. As a result, performing DFT for all generated crystals is impractical, especially because some generated samples are of low quality.

In this work, we address these two challenges jointly. To mitigate reference ambiguity, we \textit{assess stability relative to the training set}. The key intuition is that training crystals are already stable under commonly used reference sets. Therefore, if a generated crystal has formation energy comparable to that of training crystals under the same chemical system, it is also likely to be stable. This substantially reduces the dependence on an explicitly specified external reference. To avoid costly energy calculations, we further introduce an energy-aware surrogate model, \textit{\M}. Specifically, we pre-train a joint-embedding predictive architecture (JEPA)~\cite{jepa} with an InfoNCE objective~\cite{infonce} to build a crystal latent space structured by formation energy, where crystals with similar formation energies are mapped nearby and energetically dissimilar crystals are well separated. We then use \M\ embedding-based comparison as a proxy of stability assessment, enabling efficient screening of generated crystals. Building on this surrogate, we construct a simple refinement loop: we pre-train a base generative model, generate candidate crystals, select promising ones using \M, and fine-tune the base model on the selected samples. Experiments show that this screening-and-refinement pipeline substantially improves generation quality.

Our contributions are summarized as follows:
\begin{itemize}
    \item We conduct a comprehensive numerical study of the stability--novelty trade-off in de novo crystal generation, showing that it stems from the extreme precision required for crystals to remain both stable and novel.
    \item We develop \textit{\M}, an energy-aware latent surrogate for DFT-based stability evaluation, and show that it can drive a simple refinement loop that improves the generation quality.
    \item Our approach consistently outperforms strong baselines on both MP-20 and Alex-MP-20, improving up to 53.8\% and 72.7\% on \texttt{V.S.U.N} metric via DFT, respectively.
\end{itemize}

\section{Preliminaries}
\label{pre}
\paragraph{Crystal representation.}
A crystal $\mathbf{C}$ is defined by the periodic arrangement of its fundamental repeating unit, the unit cell, across the three-dimensional space. We usually describe a unit cell using three components, i.e., the atomic fractional coordinates $\bm{X}\in[0,1)^{N\times 3}$, the atom types $\bm{A}\in\mathbb{R}^{N}$, and the lattice matrix $\bm{L}\in\mathbb{R}^{3\times 3}$, where $N$ is the number of atoms in the cell. For $\bm{A}$, we consider the first 100 chemical elements and encode each atomic type with a one-hot vector, yielding $\text{one-hot}(\bm{A})\in\mathbb{R}^{N\times100}$.
To represent the lattice, we adopt the reparameterization used in prior work~\cite{mattergen}. Specifically, we factorize $\bm{L}$ through singular value decomposition and rewrite it in terms of a rotation matrix and a symmetric matrix:
\begin{equation}
    \bm{L}=\bm{U}\tilde{\bm{L}},\quad \bm{U}=\bm{W}\bm{V}^{\top},\quad \tilde{\bm{L}}=\bm{V}\bm{\Sigma}\bm{V}^{\top},
\end{equation}
where $\bm{W}$ and $\bm{V}$ denote the left and right singular matrices of $\bm{L}$, and $\bm{\Sigma}$ contains the singular values on its diagonal. Under this formulation, $\bm{U}$ corresponds to a rotation matrix, while $\tilde{\bm{L}}$ is symmetric positive definite. We then take the upper-triangular part of $\tilde{\bm{L}}$ and flatten it into a 6-dimensional vector, written as $\widehat{\bm{L}}=\mathrm{vec}(\mathrm{triu}(\tilde{\bm{L}}))\in\mathbb{R}^{6}$.
Based on these components, the $i$-th atom in crystal $\mathbf{C}$ is represented by an \textit{atom vector}:
\begin{equation}
\label{atom_rep}
    \bm{v}_i=[\bm{X}_i \, \| \, \text{one-hot}(A_i) \, \| \, \widehat{\bm{L}}]\in\mathbb{R}^{3+100+6},
\end{equation}
which concatenates its coordinate, element type, and lattice representation. The full crystal representation is then given by stacking all atom vectors:
\begin{equation}
\label{crys_rep}
\bm{V} = [\bm{v}_1, \dots, \bm{v}_N]^\top \in \mathbb{R}^{N \times (3+100+6)}.
\end{equation}

\paragraph{Thermodynamic stability.}
For a crystal $\mathbf{C}$ with $N$ atoms within unit cell, the total energy is denoted as $E_t$, and the total energy per atom is defined as $E_{t/\mathrm{atom}} = E_t/N$. 
Suppose $\mathbf{C}$ contains $k$ distinct atomic species $\{T_1,\dots,T_k\}$. Its chemical system is denoted by $T_1$--$T_2$--$\cdots$--$T_k$, and its composition is represented by $\bm{f}=(f_1,\dots,f_k)$, where $f_i\ge 0$ and $\sum_i f_i=1$. We associate $\mathbf{C}$ with an entry $\mathcal{P}=(\bm{f},E_{t/\mathrm{atom}})$ in composition--energy space.

The primary requirement of DNG is to generate crystals that are stable from a thermodynamic perspective, which is originally defined based on \emph{formation energy}. Let $\mu_i^{\mathrm{ref}}$ denote the elemental reference energy per atom of species $T_i$, typically taken from its stable elemental phase under the same computational setting. The formation energy per atom of $\mathbf{C}$ is defined as
\begin{equation}
\label{eq:formation_energy_prelim}
E_{f/atom}(\mathbf{C})
=
E_{t/\mathrm{atom}}-\sum_{i=1}^k f_i \mu_i^{\mathrm{ref}}.
\end{equation}

Given a reference dataset $\mathcal{R}$, we construct a phase diagram using all entries in $\mathcal{R}$ that belong to the same chemical system as $\mathbf{C}$ or any of its subsystems. For each reference entry $\mathcal{P}^j=(\bm{f}^j,E_{t/\mathrm{atom}}^j)$, its formation energy is $E_{f/atom}^j = E_{t/\mathrm{atom}}^j - \sum_{i=1}^k f_i^j \mu_i^{\mathrm{ref}}$.
The convex hull is then defined as the lower convex envelope in composition--formation-energy space. Accordingly, the hull formation energy at composition $\bm{f}$ is
\begin{equation}
\label{eq:formation_hull_prelim}
E_{f/atom}^{\mathrm{hull}}(\bm{f})
=
\min_{\{\lambda_j\}}
\sum_j \lambda_j E_{f/atom}^j
\quad
\text{s.t.}
\quad
\sum_j \lambda_j \bm{f}^j=\bm{f},\;\;
\sum_j \lambda_j=1,\;\;
\lambda_j\ge 0.
\end{equation}
The energy above hull is defined as
\begin{equation}
\label{eq:ehull_form_prelim}
\Delta E
=
E_{f/atom}(\mathbf{C})-E_{f/atom}^{\mathrm{hull}}(\bm{f}).
\end{equation}
In Appendix~\ref{proof}, we derive that $\Delta E$ can also be represented via total energy per atom,
\begin{equation}
\Delta E=
E_{t/\mathrm{atom}}-E_{t/\mathrm{atom}}^{\mathrm{hull}}(\bm{f}),
\label{eq:ehull_total_prelim}
\end{equation}
where $E_{t/\mathrm{atom}}^{\mathrm{hull}}(\bm{f})=\sum_j \lambda_j E_{t/\mathrm{atom}}^j$. In this work, we regard a crystal as thermodynamically stable if $\Delta E < \epsilon$, where $\epsilon=0.1$~eV/atom following prior studies~\cite{mattergen}.

\section{Stability and Novelty Trade-off: An Experimental Investigation}
\label{tradeoff}
In this section, we investigate how log-likelihood maximization relates to the evaluation criteria V.S.U.N, with a particular focus on stability and novelty.

\paragraph{Empirical demonstration of the trade-off in crystal generation.}
We reproduce multiple baselines trained on the MP-20 dataset~\cite{cdvae}, including CDVAE~\cite{cdvae}, DiffCSP~\cite{diffcsp}, DiffCSP++~\cite{diffcsppp}, FlowMM~\cite{flowmm}, FlowLLM~\cite{Flowllm}, SymmCD~\cite{symmcd}, ADiT~\cite{adit}, CrysLLMGen (7B)~\cite{crysllmgen}, SGEquiDiff~\cite{sge}, and MatterGen~\cite{mattergen}. Detailed descriptions of these models can be found in Appendix~\ref{related work}.

For evaluation, we repeat sampling 10 times and collect 1,000 crystals in each run. We then report the mean and standard deviation of \textit{stability} (S) and \textit{novelty} (N), and two combined metrics, i.e. \textit{S.U.N.} and \textit{V.S.U.N.}. In this case study, however, our main focus is on \textit{stability} and \textit{novelty}. During evaluation, structure relaxation and energy prediction are performed using MatterSim-v1-1M~\cite{mattersim}. Following prior work~\cite{flowmm}, we use MP-2023 as the reference dataset and regard crystals with energy above hull below 0.1 eV/atom as stable~\cite{mattergen}.

\begin{figure*}[t]
  \centering
  \includegraphics[scale=0.35]{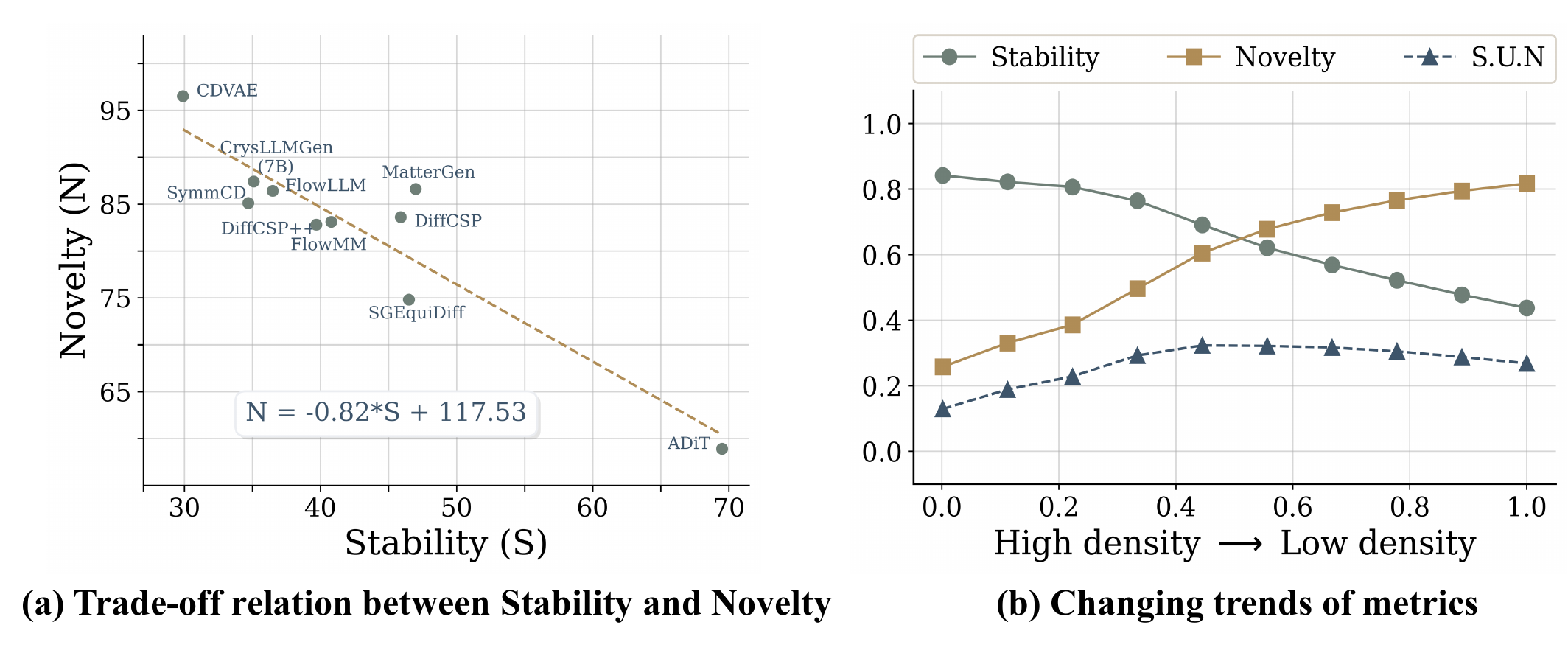}
  \caption{(a) The trade-off between stability and novelty exhibited by current crystal generative models. (b) The trends of stability, novelty, and S.U.N. as generated crystals move from regions closer to the observed distribution toward regions farther away, measured using the proxy distance defined in this section.}
  \label{case_study}
  \vskip -0.15in
\end{figure*}
The results are summarized in Table~\ref{mp20}. We further visualize the relationship between stability and novelty in Fig.~\ref{case_study}(a). CDVAE~\cite{cdvae}, for example, produces highly novel crystals, but their stability is relatively limited. In contrast, ADiT~\cite{adit} appears to stay closer to the training distribution, producing more stable but less exploratory outputs. The remaining models lie between these two extremes, showing an overall negative correlation between stability and novelty. This trend suggests that current crystal generative models struggle to balance these two objectives.

\paragraph{Why does the trade-off arise?}
Training crystals represent high-density regions within the empirical data distribution $p(x)$.
Since maximizing log-likelihood, $\log p(x)$, compels a model to prioritize these high-density regions~\cite{bishop2006pattern}, we hypothesize that the stability--novelty trade-off can be understood by examining how metrics evolve as generated samples deviate from the training distribution.

To test this, we collect 100,000 crystals generated by the aforementioned models, denoted as $\{\mathbf{C}_{gen}\}$. Because these models were optimized to maximize $\log p(x)$, the samples in $\{\mathbf{C}_{gen}\}$ are approximately drawn from the same underlying distribution as the ground-truth training set $\{\mathbf{C}_{gt}\}$. We then rank $\{\mathbf{C}_{gen}\}$ by their proximity to the high-density regions defined by $\{\mathbf{C}_{gt}\}$.

Following the \texttt{Precision} metric~\cite{cdvae}, we defined a fingerprint-based distance $\mathcal{D}$ between $\{\mathbf{C}_{gen}\}$ and $\{\mathbf{C}_{gt}\}$~\footnote{See Appendix~\ref{des_finger} for details on the fingerprint descriptors and the calculation of $\mathcal{D}$.}. As a non-learned metric, $\mathcal{D}$ provides a consistent measurement across different models and is less susceptible to the irregular behavior often seen in models processing out-of-distribution inputs. We order $\{\mathbf{C}_{\mathrm{gen}}\}$ according to $\mathcal{D}$ to observe how stability, novelty, and S.U.N. evolve across the distribution.

The resulting trends, shown in Fig.~\ref{case_study}(b), illustrate the cumulative values of each metric across percentiles of $\mathcal{D}$. As we move from regions near the observed distribution toward more distant regions, stability decreases consistently while novelty increases. Notably, we find no effective range where novelty improves substantially without a corresponding sacrifice in stability. Consequently, their intersection (S.U.N.) remains relatively stagnant. This pattern underscores a fundamental challenge in crystal generation: even minor deviations from the observed training distribution significantly degrade thermodynamic stability.

\begin{remark*}
The stability--novelty trade-off is not a new observation, as it has also been reported in recent studies~\cite{t1,t4,t3,t2}. In this section, we contribute new experimental evidence by systematically benchmarking current crystal generative models under a unified evaluation protocol. More specifically, beyond identifying the trade-off at the model level, our percentile-based analysis characterizes how stability, novelty, and S.U.N. evolve across different regions of the generated distribution. Our results show that samples farther from the empirical training distribution tend to exhibit higher novelty, but also a rapid deterioration in thermodynamic stability, thereby limiting the extent to which S.U.N. can be improved through likelihood maximization alone.
\end{remark*}

\begin{figure*}[t]
  \centering
  \includegraphics[scale=0.35]{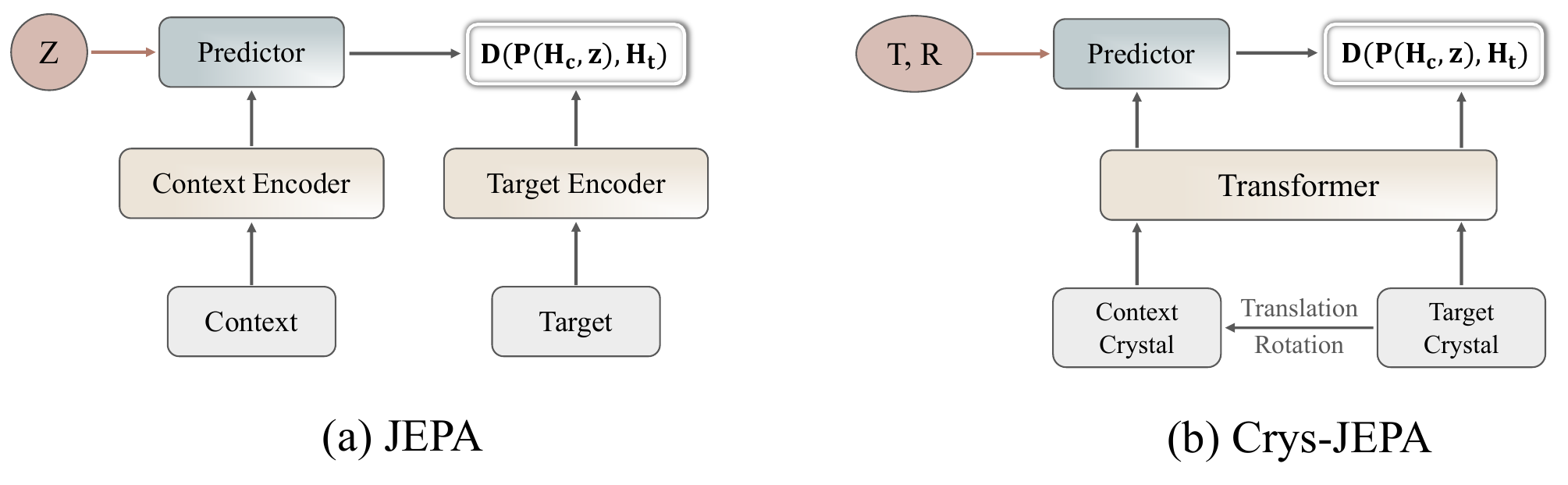}
    \vspace{-0.25cm}
  \caption{Overview of (a) JEPA architecture and (b) the proposed \M.}
  \label{framework}
  \vskip -0.15in
\end{figure*}
\section{Mitigating the Stability–Novelty Trade-off}
As shown in Fig.~\ref{case_study}(b), novelty starts from a low initial value because crystals in MP-20 are already known materials, and their local neighborhoods are therefore not novel. Without changing the likelihood-based nature of the underlying generator or the intrinsic fragility of crystals, our goal is to shift the novelty curve upward so that there exists a wider regime in which generated crystals can be both stable and novel. To this end, we screen for promising V.S.U.N. crystals among model generations and reintroduce them to refine the generator. In this section, we present \M, which serves as a practical stability surrogate during screening, together with the resulting screening-and-refinement pipeline.

\subsection{\M: Energy-aware Crystal Latent Space via JEPA}
\label{jjpp}
As discussed in Section~\ref{pre}, stability evaluation fundamentally relies on comparing the formation energies of generated crystals against those of reference crystals. In this work, we construct a unified latent space for crystals guided by formation energy per atom, $E_{f/\mathrm{atom}}$, using the JEPA framework~\cite{jepa}. The goal is to learn an embedding space in which crystals with similar formation energies are close, while crystals with larger energy differences are farther apart. The overall frameworks of JEPA and \M\ are illustrated in Fig.~\ref{framework}.

\paragraph{Context construction.}
Within JEPA, a \emph{context} is a compatible transformation of the target input that preserves its underlying semantics. While some domains provide natural context--target pairs, such as question--answer~\cite{vljepa} or text--code~\cite{llmjepa}, here we construct such pairs through data augmentation~\cite{ijepa}. Because the latent space is intended to be energy-aware, we restrict the augmentations to transformations that preserve formation energy. Specifically, we apply translation and rotation to the target crystal $\mathbf{C}_t$, rather than masking-based augmentations commonly used in vision JEPA models~\cite{ijepa, lejepa, vjepa}.

Translation acts on fractional coordinates as $\mathcal{T}(\bm{X}+\bm{t})=(\bm{X}+\bm{t})-\lfloor\bm{X}+\bm{t}\rfloor,
    \bm{t}\in[0,1)^{1\times3},$
while rotation acts on the lattice matrix as $
    \mathcal{R}(\tilde{\bm{L}})=\tilde{\bm{L}}\bm{U},
$
where $\bm{U}\in\mathrm{SO}(3)$ (special orthogonal group) is sampled from the Haar-uniform distribution, parameterized by 
$\bm{r}=(r_1,r_2,r_3) \sim \mathcal{U}([0,1]^3)$ (uniform distribution)~\cite{rotation}. The resulting context crystal $\mathbf{C}_c$ is obtained via $\mathbf{C}_c=\mathcal{R}\circ\mathcal{T}(\mathbf{C}_t)$.

\paragraph{Encoder and predictor.}
Given the augmented representations $\bm{V}_c$ and $\bm{V}_t$ as Eq.~\eqref{crys_rep}, we encode both using the same Transformer~\cite{vaswani2017attention} (refer to Appendix~\ref{model_details}). The resulting embeddings are denoted as $\bm{H}_c, \bm{H}_t \in \mathbb{R}^{N \times d}$, where $d$ represents the JEPA embedding dimension.
A predictor network is trained to infer $\bm{H}_t$ from $\bm{H}_c$ conditioned on the augmentation parameters $(\bm{t},\bm{r})$, which explicitly encode the relation between context and target. We implement the predictor as a multilayer perceptron (MLP):
\begin{equation}
    \bm{P}(\bm{H}_c,\bm{t},\bm{r})=
    \bm{W}_2 \ \sigma\!\left(\bm{H}_c+\bm{W}_1[\bm{t}\,\|\,\bm{r}]+\bm{b}_1\right)
    +\bm{b}_2,
\end{equation}
where $\{\bm{W}_1,\bm{W}_2,\bm{b}_1,\bm{b}_2\}$ are learnable parameters and $\sigma(\cdot)$ denotes \texttt{SiLU}($\cdot$) activation~\cite{silu}. 

\paragraph{Optimization.}
JEPA training aims to (i) align predicted and target embeddings, $\min D(\bm{P(H_c,t,r)}, \bm{H_t})$, and (ii) prevent representations collapse. We jointly achieve both objectives using an energy-weighted InfoNCE loss~\cite{infonce, simclr}:
\begin{equation}
\label{infonce}
\mathcal{L}=
-\frac{1}{B}\sum_i
\log
\frac{
\exp\!\left(\mathrm{sim}(\bm{P}_i,\bm{H}_t^i)/\tau\right)
}{
\sum_{k=1}^B
\exp\!\left(
\omega_{ik}\cdot \mathrm{sim}(\bm{P}_i,\bm{H}_t^k)/\tau
\right)
},
\end{equation}
where $\bm{P}_i=\bm{P}(\bm{H}_c^i,\bm{t},\bm{r})$, $\mathrm{sim}(\cdot,\cdot)$ denotes cosine similarity, $\tau>0$ is a temperature parameter, and $B$ is the batch size. The energy-aware weight $\omega_{ik}$ is defined as
\begin{equation}
\label{coe}
    \omega_{ik}=1-\exp\!\left(-\big|E_{f/\mathrm{atom}}^i-E_{f/\mathrm{atom}}^k\big|\right) \textrm{ for } k\not=i, \textrm{ and }  \ w_{ii}=1.
\end{equation}
This weighting scheme enforces stronger repulsion between embeddings of crystals with larger formation energy differences, while allowing energetically similar crystals to remain close. As a result, the learned latent space both avoids collapse and encodes a meaningful energy-aware structure, enabling embedding-based energy comparison.

\paragraph{Space visualization}
\begin{wrapfigure}[20]{r}{0.6\textwidth}
    \centering
      \includegraphics[scale=0.62]{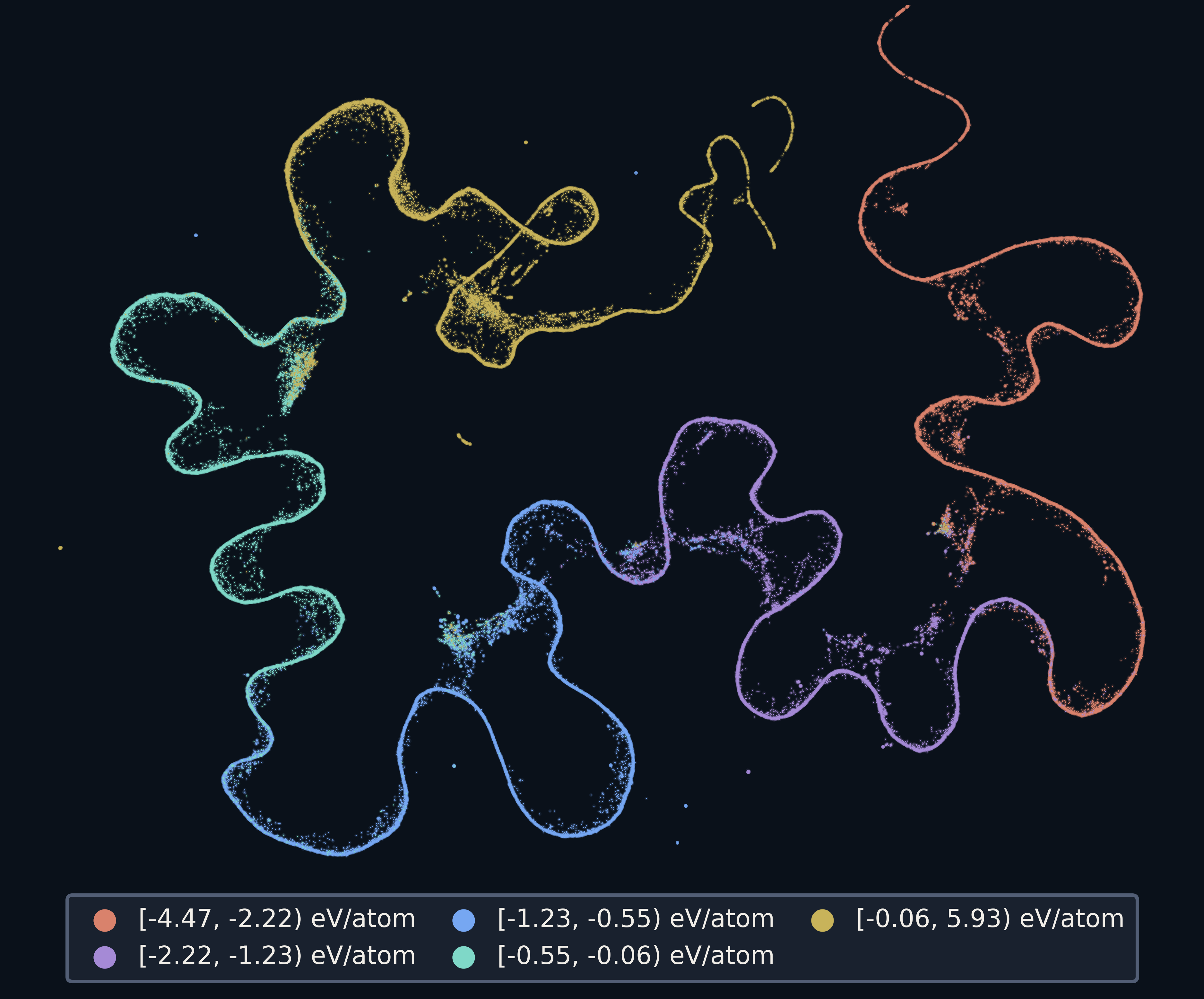}
      \caption{Visualization of the latent space learned by \M\ for 100,000 crystal structures.}
      \label{visualization}
\end{wrapfigure}
We pre-train \M\ on Material Project v.2022.10.28~\cite{MaterialsProject} and MPtrj~\cite{chgnet}. The former provides high-quality stable structures, while the latter contributes trajectory-level structurally unstable variations. Because of the resource limitation, we retain only crystals with at most 20 atoms in the unit cell.
After training, we visualize the latent space of \M\ in Fig.~\ref{visualization}. Specifically, we randomly sample 100,000 crystals from the two datasets, obtain their \M\ embeddings together with their $E_{f/\mathrm{atom}}$, and then reduce the embedding dimension using UMAP~\cite{mcinnes2018umap-software}. The samples are partitioned into five equal groups of 20,000 crystals based on $E_{f/\mathrm{atom}}$ with group specific color.

As shown in the figure, the visualized crystals form a structured manifold spanning from low $E_{f/\mathrm{atom}}$ (red) to high $E_{f/\mathrm{atom}}$ (yellow). Crystals with similar formation energies tend to be located closer together, whereas crystals with larger energy gaps tend to be more separated. This visualization provides qualitative evidence that \M\ embedding comparison can act as the surrogate of $E_{f/atom}$ difference.

\subsection{Screening-and-Refinement Pipeline}
Taking the training on MP-20, we propose the screening-and-refinement pipeline as following:
\begin{enumerate}
    \item Pre-train a target generative model $\mathcal{G}$ on MP-20.
    \item Use the model $\mathcal{G}$ to generate $\mathcal{N}$ crystals, relax them using machine learning force field, and retain those that are V.U.N relative to MP-20.
    \item For each retained crystal $\mathbf{C}$, identify its \emph{reference set}, consisting of training crystals that belong to the same chemical system as $\mathbf{C}$ or to any of its subsystems. For example, if the formula of $\mathbf{C}$ is $A_2B$, a possible reference set from the training data is $Ref=\{A, B, AB_2, A_2B_2\}$.
    \item Obtain the \M\ embedding $\bm{H}_C$ of the target crystal $\mathbf{C}$, as well as the embeddings $\{\bm{H}_i^{Ref}\}$ of the reference crystals. Based on these embeddings, we heuristically define the average distance of $\mathbf{C}$ to MP-20 as
    \begin{equation}
    D_C
    =
    \frac{1}{N_{Ref}}
    \sum_{i\in Ref}
    \left\|
    \bm{H}_C - \bm{H}_i^{Ref}
    \right\|^2,
    \label{eq:Dc}
    \end{equation}
    where $N_{Ref}$ denotes the number of candidate groups for $\mathbf{C}$.
    \item Finally, rank the retained crystals from Step 2 according to $D_C$, select top $\mathrm{k}\%$ crystals with the smallest distances, and fine-tune the base model $\mathcal{G}$ on the $\mathrm{k}\%$ crystals.
\end{enumerate}
In Step 2, we again use MatterSim-v1-1M~\cite{mattersim}, as in Section~\ref{tradeoff}. As introduced in Section~\ref{pre}, the phase diagram depends only on crystals within the same chemical system or its subsystems, which motivates the construction of the reference set in Step 3. In Step 4, instead of explicitly solving the convex-hull optimization in Eq.~\eqref{eq:formation_hull_prelim}, we treat all candidates in the reference set uniformly and use the resulting average embedding distance as a simple ranking signal. Experimental results in Section~\ref{exp} support the effectiveness of this simplification.
\begin{remark*}
Conceptually, our proposed screening-and-refinement pipeline is similar to active learning, as both aim to identify the most informative samples for model training. Similar strategies have been successfully explored in materials science, notably in GNoME~\cite{gnome} and MatterSim~\cite{mattersim}. However, traditional active learning pipelines in these works rely on computationally expensive, iterative DFT calculations to evaluate candidates and expand the training set. In contrast, our pipeline leverages \M\ as a surrogate screening signal, completely circumventing the need for further DFT annotation while maintaining effective refinement.
\end{remark*}


\subsection{Why Use Embedding-Based Comparison for Stability Screening?}
\label{why}
\begin{figure*}[h]
  \centering
  \includegraphics[scale=0.35]{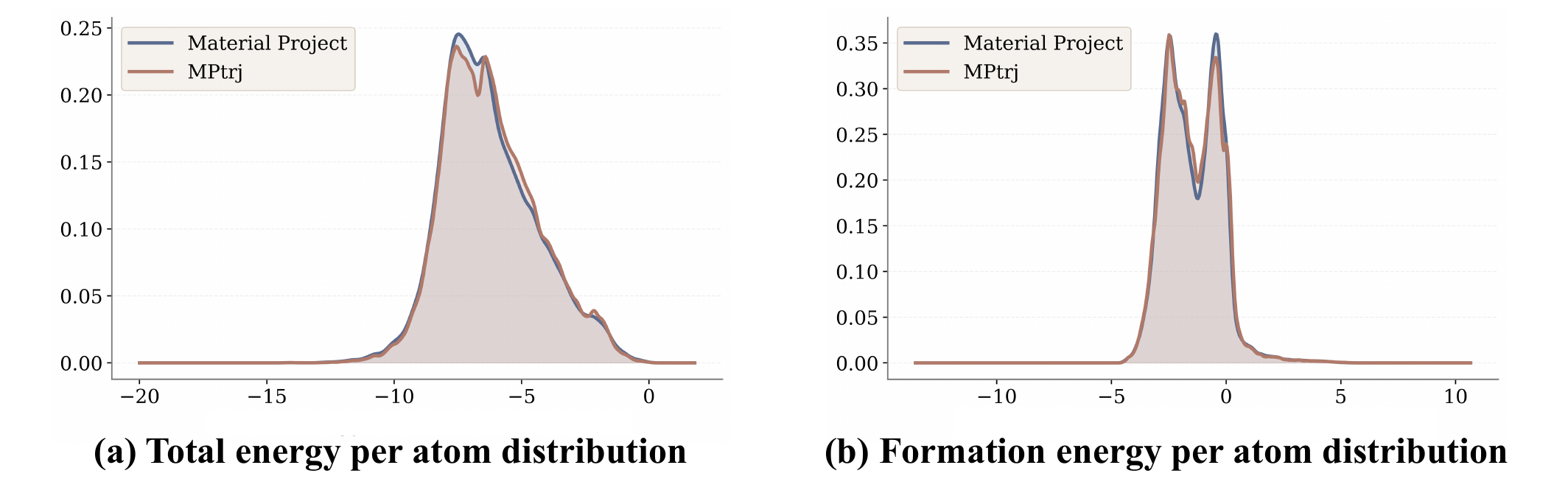}
  \caption{Distributions of $E_{f/\mathrm{atom}}$ and $E_{t/\mathrm{atom}}$ for crystals in the available datasets.}
  \label{e_distribution}
\end{figure*}
Besides \M, another option for stability screening is to use machine-learning force fields (MLFFs)~\cite{mattersim, esen}. We compare \M\ with MLFF-based screening from two perspectives:

\quad (1) \textit{Effective comparison space.} Fig.~\ref{e_distribution} shows the distributions of $E_{f/\mathrm{atom}}$ and $E_{t/\mathrm{atom}}$ in the two datasets. In both cases, most crystals are concentrated in a relatively narrow scalar range, which makes fine-grained comparison more error-prone. In contrast, \M\ performs comparison in a much broader latent space. Since our pipeline only requires relative ranking rather than absolute energy prediction, such a space can provide a more expressive signal for distinguishing candidates.
    
\quad (2) \textit{Information content.} MLFFs output scalar energy estimates (and optionally force and stress). By contrast, \M\ produces crystal embeddings that are trained to reflect formation-energy differences while still encoding structural information from the input crystal. Consequently, this approach provides a much richer informational basis for estimating thermodynamic stability than a simple scalar prediction.
    
In Section~\ref{exp}, we compare \M\ against two MLFFs, namely eSEN~\cite{esen} and CHGNet~\cite{chgnet}, and show that \M\ provides a stronger screening signal in our refinement pipeline.
\section{Numerical Experiments}
\label{exp}
In this section, we evaluate the effectiveness of \M\ and the proposed screening-and-refinement pipeline. As the target generator $\mathcal{G}$ in the pipeline, we use a basic generative model consisting of a denoising diffusion probabilistic model (DDPM)~\cite{ddpm} with a vanilla Transformer~\cite{vaswani2017attention} as the denoising network. Full model details are provided in Appendix~\ref{model_details}.

\subsection{Crystal Generation on MP-20}
\paragraph{Comparing with Generative Models.}
We first evaluate how much improvement can be brought by the proposed pipeline. We train the base model and evaluate it under the same experimental protocol as in Section~\ref{tradeoff}. The results are reported in Table~\ref{mp20}. We compare with baselines in Section~\ref{tradeoff}, and further conduct density functional theory (DFT)~\cite{perdew1996pbegga} for more accurate stability evaluation. Details are given in Appendix~\ref{dft}. Due to high cost, we only perform DFT on the 1,000 generated crystals from both the strongest baseline (MatterGen) and our model.

As shown in Table~\ref{mp20}, the base model (DDPM+Transformer) achieves higher stability but lower novelty than MatterGen, resulting in inferior overall \texttt{S.U.N.} and \texttt{V.S.U.N} performance. After applying the screening-and-refinement pipeline, most metrics improve substantially, including the most important one, \texttt{V.S.U.N}, which surpasses MatterGen.

\begin{figure*}[ht]
  \centering
  \includegraphics[scale=0.35]{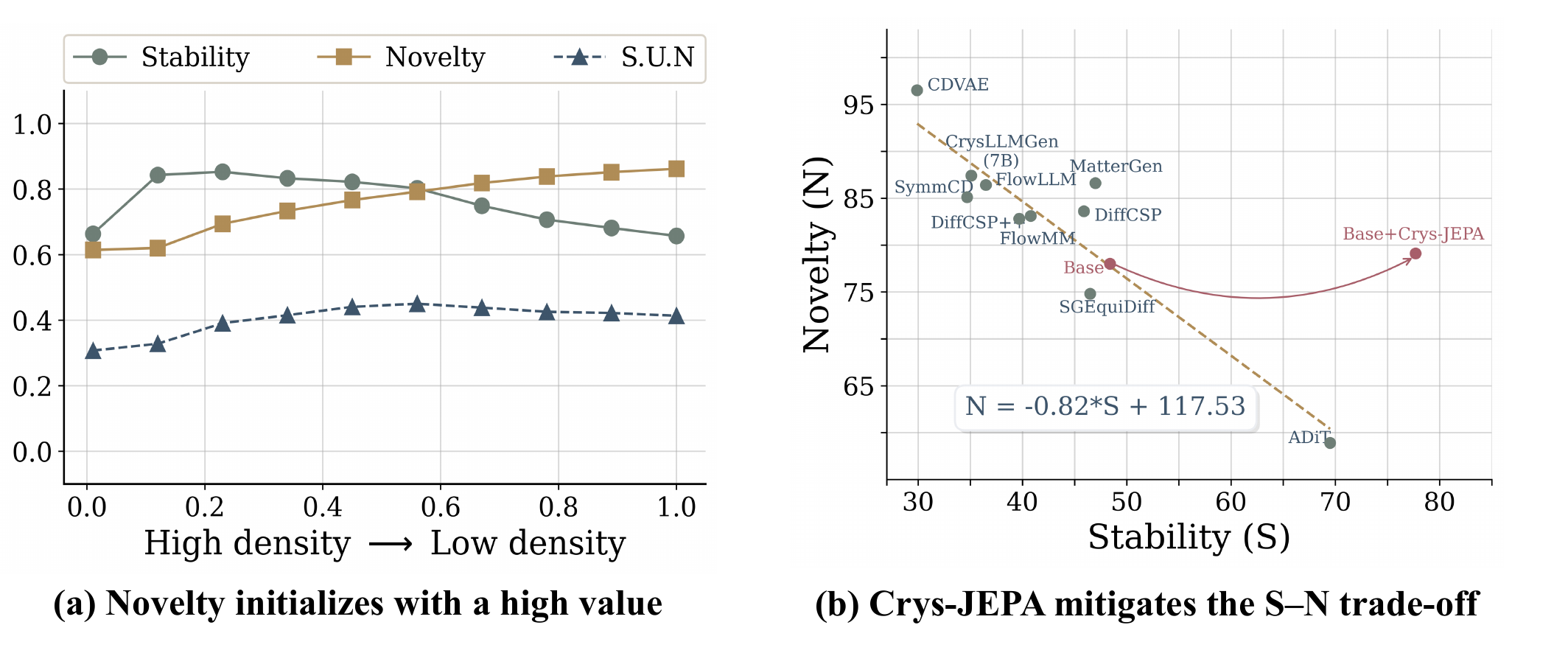}
  \caption{(a) Trends of the evaluation metrics along the proxy distance after refinement. (b) Comparison stability--novelty trade-off before and after refinement.}
  \label{addmore}
\end{figure*}

\begin{table}[t]
  \caption{Generation performance on MP-20 across 10,000 generated crystals for MLFF metrics, and with 1,000 samples used for DFT evaluation. Results are reported as the mean percentage $\pm$ standard deviation. (S: stability; N: novelty; V: validity; U: uniqueness.)}
  \vskip 0.1in
  \label{mp20}
  \small
  \centering
  \begin{tabular*}{0.93\textwidth}{@{\extracolsep{\fill}}ccc|cccc}
    \toprule[1.2pt]
           Model       &  S   &  N  & \makecell{S.U.N\\(MLFF)} & \makecell{V.S.U.N\\(MLFF)} & \makecell{S.U.N\\(DFT)} & \makecell{V.S.U.N\\(DFT)}\\
    \midrule 
CDVAE$^\dag$~\cite{cdvae} &29.9$\pm$1.2&\textbf{96.5$\pm$0.6}&27.0$\pm$1.3&22.8$\pm$1.1&-&-\\
DiffCSP~\cite{diffcsp} &45.9$\pm$1.8&83.6$\pm$0.6&30.9$\pm$1.8&25.6$\pm$1.3&-&-\\
DiffCSP++~\cite{diffcsppp} &39.7$\pm$2.0&82.8$\pm$1.0&23.8$\pm$1.3&20.3$\pm$1.2&-&-\\
FlowMM$^\dag$~\cite{flowmm} &40.8$\pm$2.0&83.1$\pm$1.1&25.3$\pm$1.6&21.0$\pm$1.3&-&-\\
FlowLLM$^\dag$~\cite{Flowllm} &36.5$\pm$1.5&86.4$\pm$1.5&25.1$\pm$0.6&21.3$\pm$0.8&-&-\\
SymmCD$^\dag$~\cite{symmcd} &34.7$\pm$1.4&85.1$\pm$1.5&19.0$\pm$0.9&15.7$\pm$0.8&-&-\\
ADiT~\cite{adit} &69.5$\pm$1.1&58.9$\pm$0.9&30.3$\pm$0.8&27.2$\pm$1.3&-&-\\
CrysLLMGen (7B)$^\dag$~\cite{crysllmgen} &35.1$\pm$1.9&87.4$\pm$1.1&22.9$\pm$0.9&20.4$\pm$0.9&-&-\\
SGEquiDiff~\cite{sge} &46.5$\pm$1.9&74.8$\pm$1.2&23.5$\pm$0.9&20.0$\pm$1.0&-&-\\
MatterGen~\cite{mattergen} &47.0$\pm$1.1&86.6$\pm$1.0&34.6$\pm$1.2&29.2$\pm$0.9&30.8&26.4\\
    \midrule
    Base model &48.4$\pm$1.4&78.0$\pm$1.6&27.8$\pm$1.1&23.3$\pm$1.3&-&-\\
    w/ \M\ &\textbf{76.6$\pm$1.1}&83.3$\pm$1.1&\textbf{45.2$\pm$1.4}&\textbf{44.9$\pm$1.4}&\textbf{48.1}&\textbf{47.9}\\
    \bottomrule[1.2pt]
  \end{tabular*}
\end{table}
We then collect all 10,000 crystals generated by the base model and repeat the analysis in Fig.~\ref{case_study}(b). The resulting trends are shown in Fig.~\ref{addmore}(a). Compared with Fig.~\ref{case_study}(b), the novelty curve starts from a noticeably higher value after refinement. In addition, there exists a range, approximately $(0, 0.5)$, in which stability remains high while novelty continues to increase. This regime allows the model to maintain both stability and novelty simultaneously, thereby sustaining a higher \texttt{V.S.U.N}. As in Fig.~\ref{addmore}(b), the proposed pipeline actually pushes the base model beyond the stability-novelty trade-off.


\paragraph{Comparing \M\ with fingerprints and MLFFs}
In this experiment, we compare \M\ with two MLFFs, eSEN~\cite{esen} and CHGNet~\cite{chgnet}. Specifically, we use each MLFF to predict crystal energies, compute $\Delta E$ for generated crystals relative to the training set, and rank the generated candidates by $\Delta E$ before selection. For completeness, we also consider the fingerprint-based distance introduced in Section~\ref{tradeoff} as an alternative ranking signal at Step 4 of the pipeline. The results are summarized in Table~\ref{mp20_mlff}. Besides generation metrics, we also report the training-set size used by each model and the corresponding inference cost. To ensure a fair runtime comparison and eliminate the effect of batch size, we disable batched inference by setting the batch size to 1. Each model sequentially processes 10,000 crystals, and the total inference time is reported.
\begin{table}[t]
  \caption{Comparison of different screening ways on MP-20, including fingerprint-based ranking, MLFF-based ranking, and the proposed \M. Results are averaged over 10,000 sampled crystals and reported as mean percentage $\pm$ standard deviation.} 
  \vskip 0.1in
  \label{mp20_mlff}
  \small
  \centering
  \renewcommand\arraystretch{1.2}
  \resizebox{1\columnwidth}{!}
{
  \begin{tabular}{cccc|cccc}
    \toprule[1.2pt]
           \%       &  V & S   &  N  &  S.U.N &  V.S.U.N & \makecell{Inference Time (s)\\10,000 Crystals} & \makecell{\# Training\\Samples}\\
    \midrule 
    Base model &82.1$\pm$1.5&48.4$\pm$1.4&78.0$\pm$1.6&27.8$\pm$1.1&23.3$\pm$1.3&-&27,136\\
    \midrule
    w/ Fingerprints &94.1$\pm$1.0&53.1$\pm$1.4&72.9$\pm$1.2&22.9$\pm$0.7&21.9$\pm$0.7&1651.69&-\\
    w/ CHGNet~\cite{chgnet} &96.8$\pm$0.3&74.7$\pm$1.3&77.4$\pm$0.6&39.8$\pm$0.9&38.2$\pm$0.7&339.49&1,580,395\\
    w/ eSEN-30M-MP~\cite{esen} &95.8$\pm$0.3&\textbf{77.2$\pm$1.0}&80.2$\pm$0.9&44.0$\pm$0.8&41.0$\pm$1.1&843.53&1,580,395\\ 
    \midrule
    w/ \M &\textbf{99.0$\pm$0.3}&76.6$\pm$1.1&\textbf{83.3$\pm$1.1}&\textbf{45.2$\pm$1.4}&\textbf{44.9$\pm$1.4}&\textbf{48.87}&\textbf{839,568}\\
    \bottomrule[1.2pt]
  \end{tabular}
  }
\end{table}
\begin{enumerate}
    \item Fingerprints capture structural and compositional similarity, but they are not directly energy-aware. Consequently, they are less reliable for screening stable crystals.

\item \M\ substantially improves the validity of the base model. This is because the \M\ embedding also encodes structural information. When \M\ is used to screen generated crystals, the selected candidates are more likely to be structurally close to the valid training crystals.

\item eSEN is a strong baseline for filtering high-quality crystals. However, it explicitly models all interatomic interactions within a cutoff radius, which leads to a high inference cost. In contrast, \M\ adopts a vanilla Transformer encoder and focuses only on atoms within the unit cell, resulting in a much more efficient screening procedure.

\item Both MLFFs are trained on the full MPTrj dataset with force, stress, and energy labels. By contrast, as described in Section~\ref{jjpp}, \M\ uses roughly half of the crystals and only energy labels. This suggests that scaling \M\ with richer supervision and more training data could further improve its performance.

\item A key requirement for computing $\Delta E$ via MLFFs is the reference crystals (in this case, the training set) must have DFT-calculated energies for an accurate convex hull. In contrast, our \M\ solely relies on crystal structures and a simple Euclidean metric $D_C$ per Eq.~\ref{eq:Dc}.
\end{enumerate}

\subsection{Crystal Generation base on Alex-MP-20}
\begin{table}[ht]
  \caption{Evaluation of 10,000 crystals generated by models trained on Alex-MP-20. 
  }
  \vskip 0.1in
  \label{alex}
  \small
  \centering
  \begin{tabular*}{0.94\textwidth}{@{\extracolsep{\fill}}ccc|cccc}
    \toprule[1.2pt]
           Model       &  S   &  N  &  \makecell{S.U.N\\(MLFF)} & \makecell{V.S.U.N\\(MLFF)} & \makecell{S.U.N\\(DFT)} & \makecell{V.S.U.N\\(DFT)}\\
    \midrule 
    MatterGen &73.1$\pm$1.0&67.9$\pm$1.4&43.5$\pm$1.2&37.4$\pm$1.4&40.1&35.1\\
    \midrule
    Base model &69.0$\pm$0.9&61.8$\pm$1.2&33.2$\pm$1.5&28.8$\pm$1.0&-&-\\
    w/ \M\ &\textbf{78.9$\pm$1.1}&\textbf{85.1$\pm$0.7}&\textbf{64.7$\pm$0.9}&\textbf{64.6$\pm$0.9}&\textbf{64.1}&\textbf{64.1}\\
    \bottomrule[1.2pt]
  \end{tabular*}
\end{table}
We further verify the effectiveness of \M\ and the screening-and-refinement pipeline on the Alex-MP-20 dataset~\cite{mattergen}, using the \texttt{MP2020correction} reference following~\cite{mattergen}. The results are presented in Table~\ref{alex}. Once again, the proposed pipeline yields a substantial improvement in metrics such as \texttt{V.S.U.N.}. In particular, compared to MatterGen, \texttt{V.S.U.N.} increases by 82.6\% via DFT.

\section{Conclusion}
\label{conclusion}
In this paper, we provide a numerical investigation of the stability–novelty trade-off in de novo crystal generation and propose \M, an energy-aware embedding model for efficient stability screening. By mapping generated crystals alongside known training references within a shared learned latent space, \M\ enables a screening-and-refinement pipeline that identifies promising candidates and helps move beyond this inherent trade-off.

\paragraph{Limitations and broader impact.}
While \M\ was trained on a relatively limited dataset, its performance could be further enhanced through larger-scale data and advanced architectural refinements. Although this work significantly reduces the computational cost of materials screening, candidates selected by the surrogate model should be rigorously validated via first-principles methods before being considered definitive material discoveries.


\begin{ack}
XB is supported by MOE AcRF T1 Grant ID 251RES2423 and NRF AI4SCT Grant ID 20250024. KSN acknowledges support by the National Research Foundation, Singapore under its AI Singapore Programme (AISG Award No: AISG3-RP-2022-028), by the Ministry of Education, Singapore under Research Centre of Excellence award to the Institute for Functional Intelligent Materials, I-FIM (project No. EDUNC-33-18-279-V12), 
by the MAT-GDT Program at A*STAR via the AME Programmatic Fund by the Agency for Science, Technology and Research under Grant No. M24N4b0034 
and by the Tier 3 program (MOE-MOET32024-0001). The computational work for this article was partially performed on resources of the National Supercomputing Centre, Singapore (https://www.nscc.sg).
\end{ack}

\newpage
\bibliographystyle{plain}
\bibliography{bib}

\newpage
\appendix
\section{Thermodynamic Stability Calculation}
\label{proof}
In this section, we give a brief derivation that energy above hull can also be measured using total energy per atom. Firstly, we rewrite Eq.~\eqref{eq:formation_hull_prelim} and Eq.~\ref{eq:ehull_form_prelim} here,

\begin{subequations}
\label{start}
\begin{align}
E_{f/atom}^{\mathrm{hull}}(\bm{f})
&=
\min_{\{\lambda_j\}}
\sum_j \lambda_j E_{f/atom}^j
\quad
\text{s.t.}
\quad
\sum_j \lambda_j \bm{f}^j=\bm{f},\;\;
\sum_j \lambda_j=1,\;\;
\lambda_j\ge 0,
\label{eq:1a}\\
\Delta E
&=
E_{f/atom}(\mathbf{C})-E_{f/atom}^{\mathrm{hull}}(\bm{f}). \label{eq:1b}
\end{align}
\end{subequations}
Then, substituting the $E_{f/atom}^j = E_{t/\mathrm{atom}}^j - \sum_{i=1}^k f_i^j \mu_i^{\mathrm{ref}}$ into Eq.~\eqref{eq:1a} gives
\begin{equation}
\begin{aligned}
E_{f/atom}^{\mathrm{hull}}(\bm{f})
&=
\min_{\{\lambda_j\}}
\sum_j \lambda_j
\left(
E_{t/\mathrm{atom}}^j - \sum_{i=1}^k f_i^j \mu_i^{\mathrm{ref}}
\right) \\
&=
\min_{\{\lambda_j\}}
\left[
\sum_j \lambda_j E_{t/\mathrm{atom}}^j
-
\sum_{i=1}^k
\left(\sum_j \lambda_j f_i^j\right)\mu_i^{\mathrm{ref}}
\right].
\end{aligned}
\end{equation}
Using the composition-matching constraint $\sum_j \lambda_j \bm{f}^j=\bm{f}$ in Eq.~\eqref{eq:formation_hull_prelim}, we have $\sum_j \lambda_j \bm{f}_i^j=\bm{f}_i$ for all $i$, and thus
\begin{equation}
E_{f/atom}^{\mathrm{hull}}(\bm{f})
=
E_{t/atom}^{\mathrm{hull}}(\bm{f})-\sum_{i=1}^k f_i \mu_i^{\mathrm{ref}},
\end{equation}
where
\begin{equation}
E_{t/atom}^{\mathrm{hull}}(\bm{f})
=
\min_{\{\lambda_j\}}
\sum_j \lambda_j E_{t/\mathrm{atom}}^j
\quad
\text{s.t.}
\quad
\sum_j \lambda_j \bm{f}^j=\bm{f},\;\;
\sum_j \lambda_j=1,\;\;
\lambda_j\ge 0
\end{equation}
is the convex hull energy in composition--total-energy space. Combining this with Eq.~\eqref{eq:1b}, the elemental reference term cancels:
\begin{equation}
\begin{aligned}
\Delta E
&=
\left(E_{t/\mathrm{atom}}-\sum_{i=1}^k f_i \mu_i^{\mathrm{ref}}\right)
-
\left(E_{t/atom}^{\mathrm{hull}}(\bm{f})-\sum_{i=1}^k f_i \mu_i^{\mathrm{ref}}\right) \\
&=
E_{t/\mathrm{atom}}-E_{t/atom}^{\mathrm{hull}}(\bm{f}).
\end{aligned}
\end{equation}
Therefore, although the phase diagram is originally defined using formation energies, the energy above hull can be computed equivalently as the difference between the crystal total energy per atom and the convex hull energy at the same composition. This $\Delta E$ computing has been adopted by \texttt{pymatgen}~\cite{ong2013python} internally.
\section{Model Details}
\label{model_details}
\subsection{Diffusion-Based Generation}
Diffusion models~\cite{origin} have recently become a leading framework for generative modeling. In this work, we adopt the denoising diffusion probabilistic model (DDPM)~\cite{ddpm} as our generative backbone.

DDPM consists of a forward noising process and a learned reverse denoising process. In the forward process, Gaussian noise is gradually added to a clean sample $\bm{x}_0$ over $T$ timesteps:
\begin{equation}
\bm{x}_t = \sqrt{1-\beta_t}\,\bm{x}_{t-1} + \sqrt{\beta_t}\,\bm{\epsilon}_t
= \sqrt{\bar{\alpha}_t}\,\bm{x}_0 + \sqrt{1-\bar{\alpha}_t}\,\bm{\epsilon}, 
\quad \bm{\epsilon} \sim \mathcal{N}(0, \bm{I}),
\end{equation}
where $\{\beta_t\}_{t=1}^T$ denotes a predefined variance schedule with $\beta_1 < \cdots < \beta_T$, $\alpha_t := 1 - \beta_t$, and $\bar{\alpha}_t := \prod_{s=1}^t \alpha_s$. This formulation yields a closed-form expression for the conditional distribution $q(\bm{x}_t \mid \bm{x}_0)$.

The reverse process reconstructs $\bm{x}_0$ from noise using a parameterized model. Specifically, a neural network $\bm{\epsilon}_\theta(\bm{x}_t, t)$ is trained to estimate the injected noise at each timestep. The corresponding training objective is:
\begin{equation}
\mathcal{L}_{\text{DDPM}}(\theta)
= \mathbb{E}_{t, \bm{x}_0, \bm{\epsilon}}
\left[
\left\|
\bm{\epsilon} - \bm{\epsilon}_\theta(\bm{x}_t, t)
\right\|_2^2
\right],
\end{equation}
where $t$ is sampled uniformly from $\{1, \ldots, T\}$.

During sampling, generation starts from Gaussian noise $\bm{x}_T \sim \mathcal{N}(0, \bm{I})$ and iteratively applies the learned reverse transitions to recover $\bm{x}_0$. The reverse transition is modeled as:
\begin{equation}
q(\bm{x}_{t-1} \mid \bm{x}_t, \bm{x}_0)
= \mathcal{N}\bigl(\bm{x}_{t-1}; \tilde{\bm{\mu}}_t(\bm{x}_t, \bm{x}_0), \tilde{\beta}_t \bm{I}\bigr),
\end{equation}
with
\begin{equation}
\tilde{\bm{\mu}}_t =
\frac{1}{\sqrt{\alpha_t}}
\left(
\bm{x}_t -
\frac{\beta_t}{\sqrt{1-\bar{\alpha}_t}}
\bm{\epsilon}_\theta(\bm{x}_t, t)
\right), 
\quad
\tilde{\beta}_t =
\frac{1-\bar{\alpha}_{t-1}}{1-\bar{\alpha}_t}\beta_t.
\end{equation}

\subsection{Transformer as Backbones}
\begin{wrapfigure}[20]{r}{0.6\textwidth}
    \centering
      \includegraphics[scale=0.3]{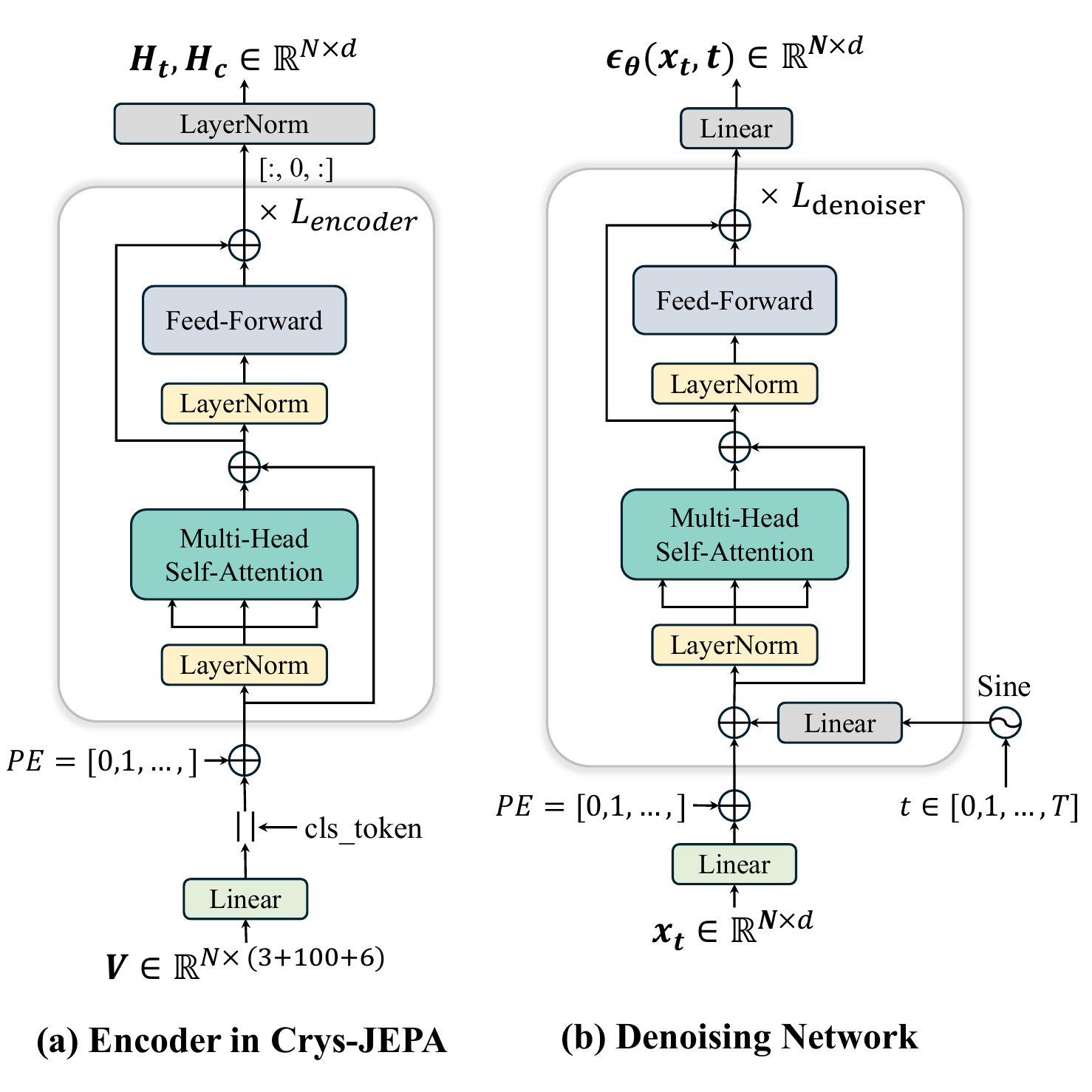}
      \caption{Two backbone Transformers.}
      \label{transs}
\end{wrapfigure}
Our framework employs two Transformer-based architectures: a crystal encoder in \M\ and a denoising network in DDPM. The detailed designs are shown in Fig.~\ref{transs}. Both are built upon the standard Transformer~\cite{vaswani2017attention}, with several practical modifications.

\paragraph{Pre-normalization.}
We adopt pre-normalization before both the self-attention and feed-forward layers, following~\cite{pre-norm}, to improve optimization stability.

\paragraph{Global Representation.}
To obtain a global crystal representation in Crys-JEPA, we introduce a learnable \texttt{[CLS]} token~\cite{vit}. This token is concatenated with atom embeddings and processed through the Transformer. Its final output serves as the global embedding.

\paragraph{Timestep Conditioning.}
In the DDPM denoising network, the timestep $t$ is encoded using sinusoidal embeddings followed by a linear projection. The resulting representation is added to the input features at each Transformer layer.

\subsection{Hyperparameter Settings}
For the Crys-JEPA encoder, we use an 8-layer Transformer with hidden dimension 512 and 16 attention heads, without dropout.

For the diffusion model, we set the number of timesteps to $T = 256$ and adopt a cosine noise schedule~\cite{Improved}. The denoising Transformer consists of 12 layers with hidden dimension 1024 and 8 attention heads, and a dropout rate of 0.01.

\section{More Experiments}
\subsection{Hyper-Parameter Sensitivity Analysis}
\begin{figure}[t]
    \centering
    \begin{minipage}{0.3\linewidth}
        \centering
        \includegraphics[width=\linewidth]{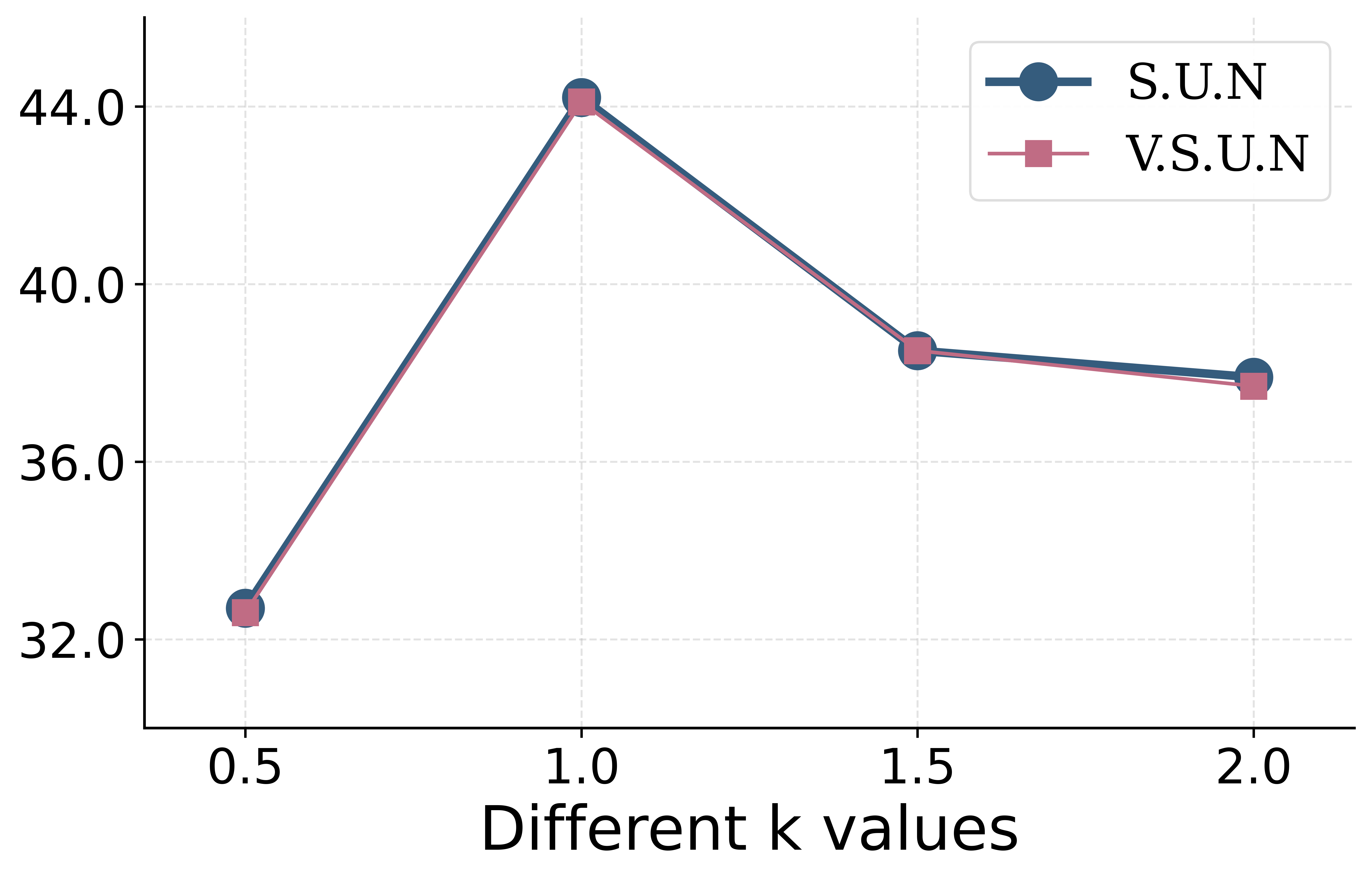}
        \caption*{(a) $k$ on MP-20}
    \end{minipage}
    \begin{minipage}{0.3\linewidth}
        \centering
        \includegraphics[width=\linewidth]{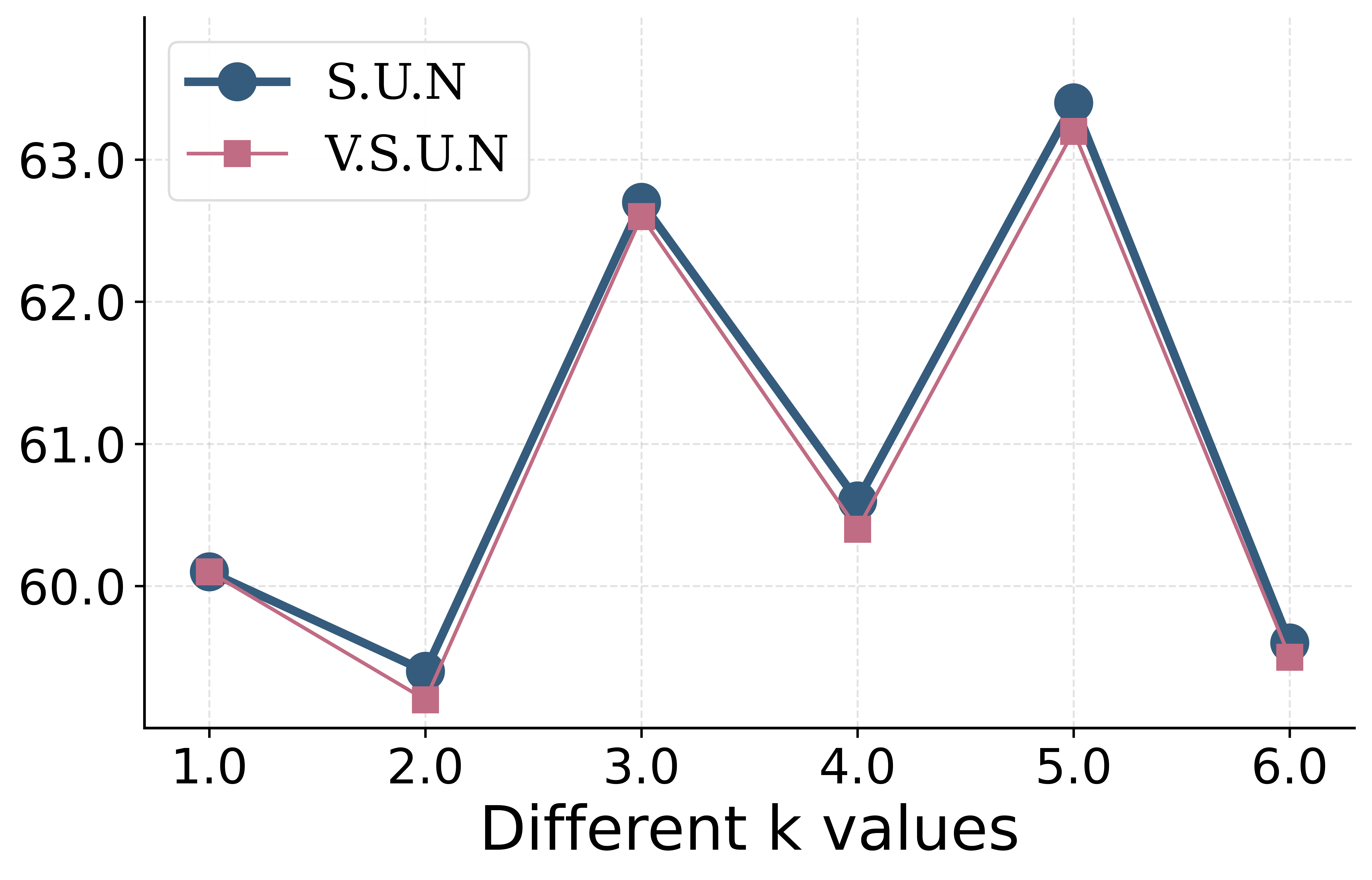}
        \caption*{(b) $k$ on Alex-MP-20}
    \end{minipage}
    \begin{minipage}{0.3\linewidth}
        \centering
        \includegraphics[width=\linewidth]{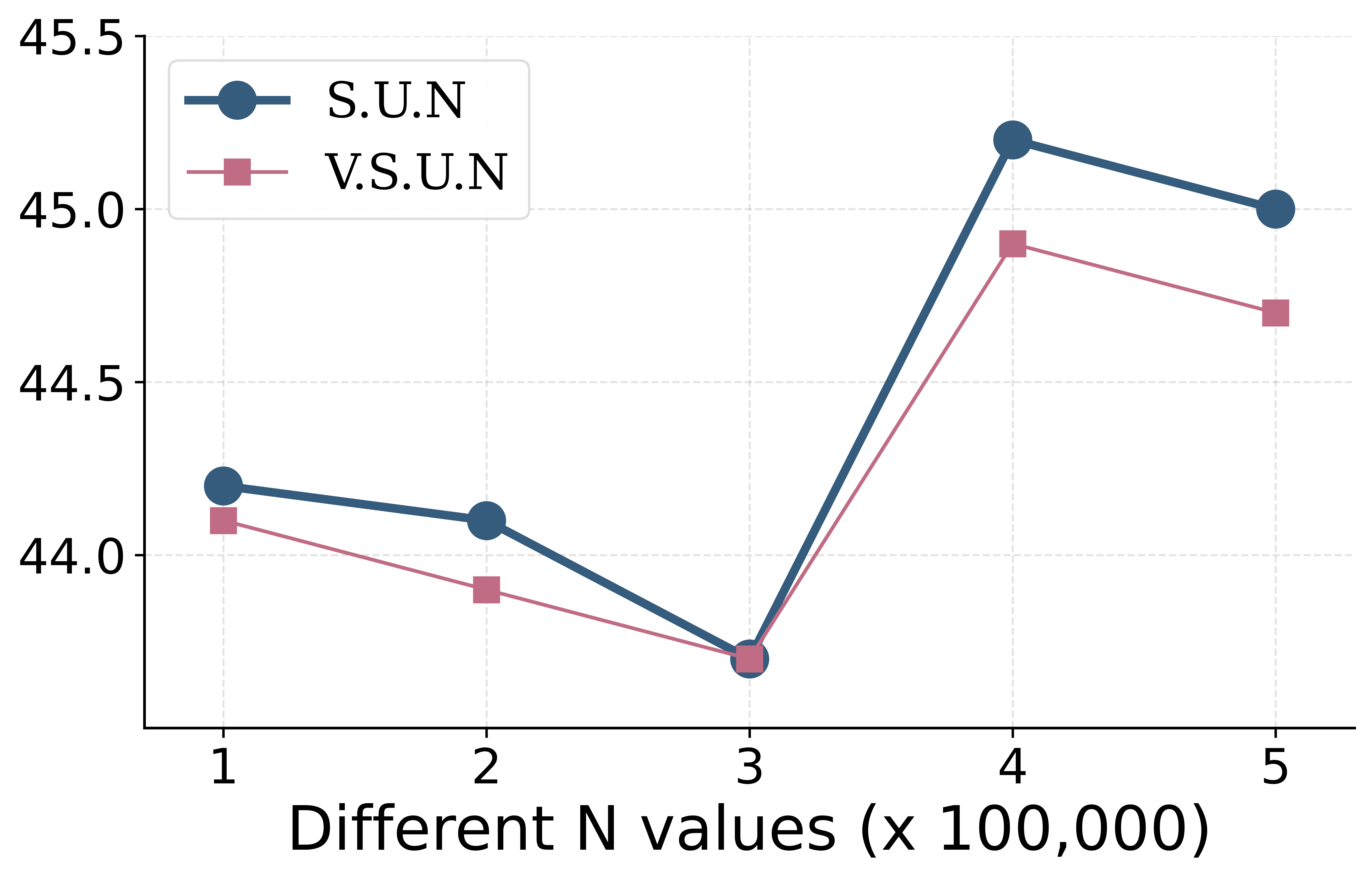}
        \caption*{(c) $\mathcal{N}$ on MP-20}
    \end{minipage}
    \caption{Test the changing trend of \texttt{S.U.N} and \texttt{V.S.U.N} with the increasing of $k$ or $\mathcal{N}$.}
    \label{k_n}
\end{figure}

In the screening-and-refinement pipeline, two key hyper-parameters, namely $\mathrm{k}$ and $\mathcal{N}$, play important roles. The parameter $\mathrm{k}$ governs the number of the selected crystals at step 5 for fine-tuning, while $\mathcal{N}$ determines the number of initial crystals at step 1. The results are visually presented in Fig.~\ref{k_n}. With increasing $\mathrm{k}$, more and more crystals distant from the training set will be included into the fine-tuning. This is a double-edged sword, bringing more various crystals while decreasing the stability. Consequently, the \texttt{S.U.N} and \texttt{V.S.U.N} generally rises and then falls. For Increasing $N$ generally improves both \texttt{S.U.N} and \texttt{V.S.U.N}, indicating that a larger candidate pool provides more high-quality samples for subsequent screening. This observation is consistent with the intuition that the screening-and-refinement pipeline benefits from a richer pool, as it increases the probability of discovering crystals that simultaneously satisfy stability, uniqueness, and novelty.

\subsection{Further Analysis of \M\ Representation}
\paragraph{Energy-aware JEPA representation.}
The main \M\ model is trained with the energy-aware InfoNCE objective in Eq.~\eqref{infonce}, where the weighting term in Eq.~\eqref{coe} modulates the repulsion between crystal embeddings according to their formation-energy differences. This design aims to organize the latent space such that crystals with similar formation energies are mapped close to each other, while crystals with larger formation-energy gaps are pushed farther apart. As shown in the main text, this representation can serve as an effective surrogate for stability screening in the proposed screening-and-refinement pipeline.

\begin{figure*}[ht]
  \centering
  \includegraphics[scale=0.4]{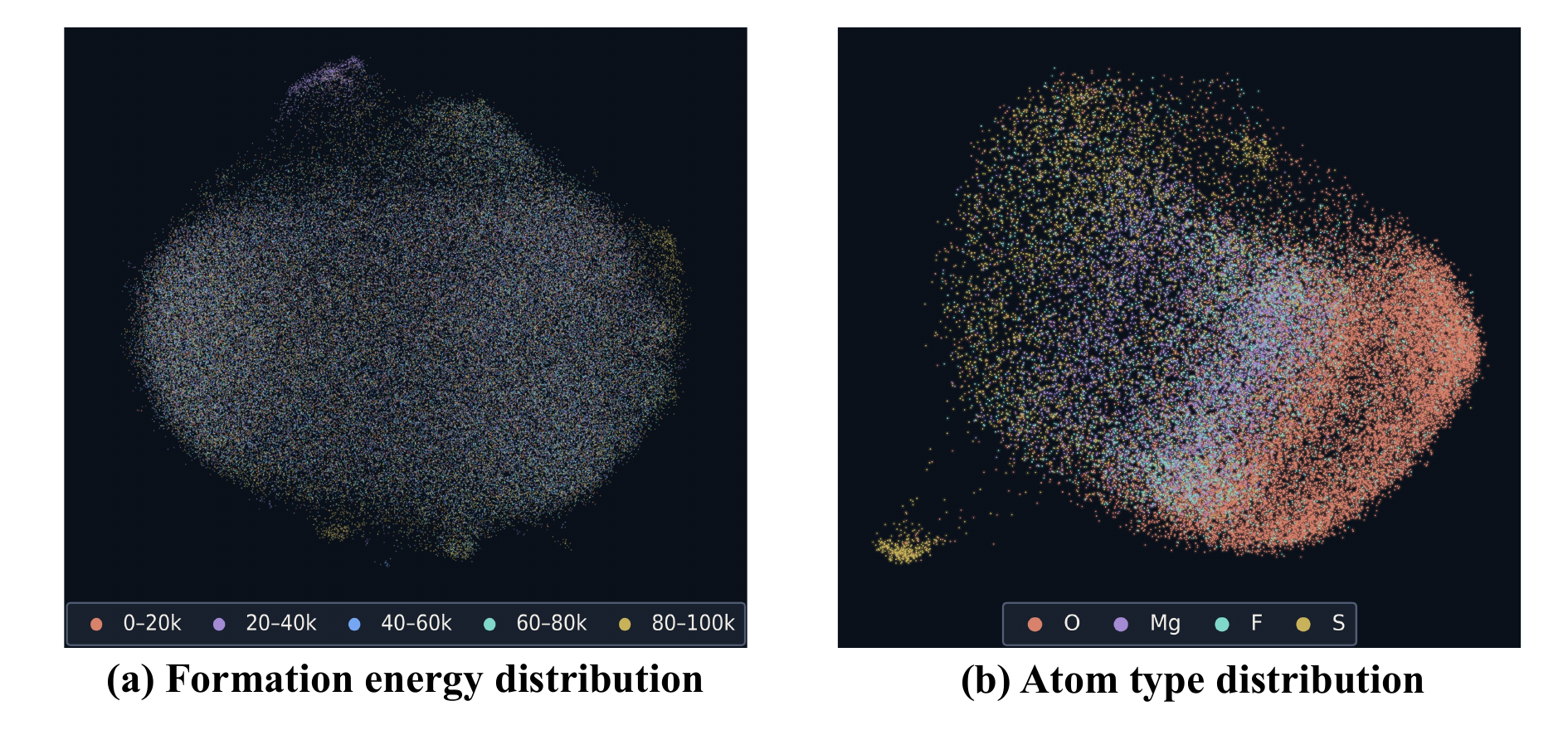}
  \caption{(a) Visualization of the latent space learned by \M\ without $\omega_{ik}$. (b) Distribution of top four most frequent atom types.}
  \label{wo_coe}
  \vskip -0.15in
\end{figure*}
To further understand the learned representation, we analyze the latent space under a related setting where the energy-aware weighting term is removed and all negative samples are treated equally. The resulting latent space is visualized in Fig.~\ref{wo_coe}(a). Compared with the energy-aware representation, the embeddings no longer exhibit a clear organization with respect to formation energy. Instead, the model appears to capture more generic structural and compositional patterns induced by the augmentation-based JEPA objective.

We further inspect atom-level embeddings by visualizing several frequent elements, including O, Mg, F, and S, as shown in Fig.~\ref{wo_coe}(b). The embeddings show partial separation among different element types, suggesting that the model still retains chemically meaningful information even without explicit energy-aware weighting. This observation is consistent with the intuition that translation and rotation augmentations encourage the encoder to preserve intrinsic crystal information.

\paragraph{Limitation.}
However, these results should be interpreted with caution. Although the visualizations suggest that the learned embeddings contain certain structural, compositional, or chemical signals, they do not by themselves provide conclusive evidence that the representation is globally well-structured. In particular, JEPA-style objectives may suffer from representation collapse or partial collapse, where embeddings occupy a low-dimensional subspace or concentrate around a limited set of directions. In such cases, low-dimensional visualization methods such as UMAP may still produce visually separable clusters, but the apparent structure may not faithfully reflect a robust or uniformly informative latent space.

This issue is especially important for our setting because \M\ is used as a ranking signal for stability screening. The screening pipeline requires not only that the representation separates some obvious chemical patterns, but also that distances in the embedding space provide a reliable and fine-grained comparison between generated crystals and reference crystals. Therefore, while the above analysis provides useful qualitative evidence, it remains insufficient for establishing that the learned representation is free from collapse or that all embedding dimensions contribute meaningfully to the energy-aware comparison.

\begin{wrapfigure}[22]{r}{0.6\textwidth}
\vspace{-1.em}
    \centering
      \includegraphics[scale=0.62]{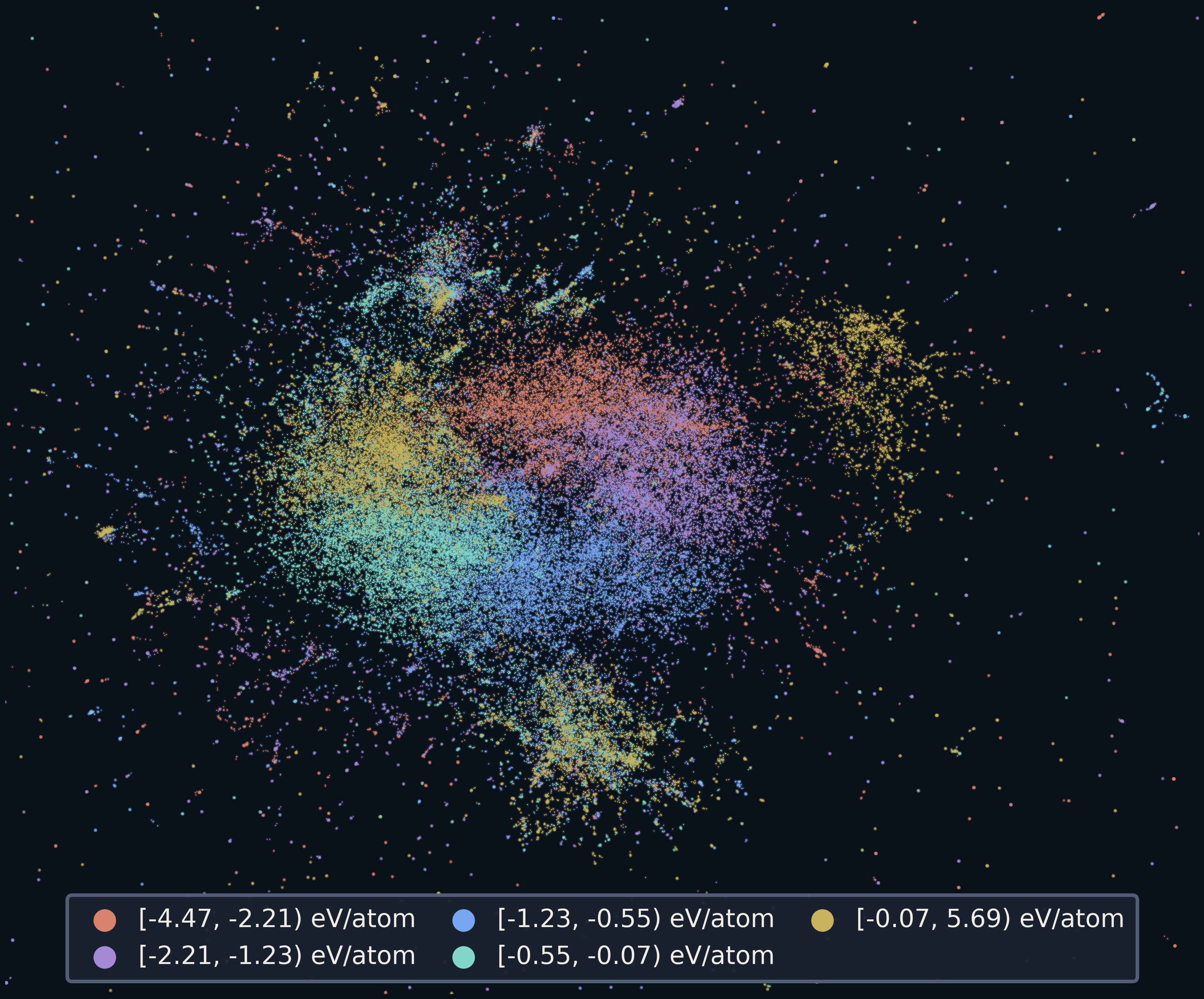}
      \caption{Visualization of the latent space learned by \M\ with SIGReg and MSE loss. Although formation-energy labels are not used during training, the learned latent space still exhibits a partial organization with respect to formation energy.}
      \label{jepa_sig}
\end{wrapfigure}
\paragraph{Collapse-resistant JEPA with SIGReg and MSE loss.}
Motivated by this limitation, we further explore an alternative JEPA training objective based on SIGReg~\cite{lejepa}. SIGReg is a recently proposed regularization method for preventing representation collapse in joint-embedding predictive architectures. Instead of relying on negative samples, SIGReg projects a batch of embeddings onto multiple random one-dimensional directions and encourages each projected distribution to match a standard Gaussian. By enforcing Gaussianity across random projections, the batch-level embedding distribution is encouraged to approximate an isotropic Gaussian. This regularization helps prevent trivial collapse and encourages information to be spread across multiple embedding dimensions.

In this variant, we replace the energy-aware InfoNCE objective in Eq.~\eqref{infonce} with a combination of mean squared error (MSE) alignment and SIGReg regularization. The MSE term aligns the predicted target embedding with the encoded target embedding:
\[
\mathcal{L}_{\mathrm{MSE}}
=
\left\|
P(H_c, t, r) - H_t
\right\|_2^2,
\]
where \(H_c\) and \(H_t\) denote the context and target embeddings, respectively, and \(P(\cdot)\) is the predictor conditioned on the augmentation parameters. The full objective is then defined as
\[
\mathcal{L}
=
\mathcal{L}_{\mathrm{MSE}}
+
\lambda_{\mathrm{SIG}}
\mathcal{L}_{\mathrm{SIGReg}},
\]
where \(\lambda_{\mathrm{SIG}}\) controls the strength of the SIGReg regularization.

Unlike the energy-aware InfoNCE objective used in the main model, this variant does not use formation-energy labels. Therefore, it provides a fully self-supervised JEPA training paradigm based only on augmentation consistency and collapse prevention. The resulting latent space is visualized in Fig.~\ref{jepa_sig}. Interestingly, although formation energy is not explicitly used during training, the learned embeddings still exhibit a partial organization with respect to formation energy. This suggests that energy-related information may be partially recoverable from the structural and compositional patterns preserved by the self-supervised objective.

Nevertheless, we view this SIGReg-based variant as complementary to, rather than a replacement for, the energy-aware \M\ objective used in the main experiments. The energy-aware objective directly injects formation-energy differences into the latent geometry and is therefore better aligned with our stability-screening goal. By contrast, SIGReg and MSE provide a promising collapse-resistant representation learning framework, but currently lack an explicit mechanism for ordering crystals by formation-energy differences. Future work will investigate how SIGReg-style regularization can be combined with energy supervision to obtain a latent space that is both collapse-resistant and thermodynamically informative.

\section{Experimental Details}
\subsection{Datasets Descriptions}
We consider the following two datasets in the numerical experiments:
\begin{itemize}
    \item \textbf{MP-20} is a realistic benchmark curated from the Materials Project~\cite{MaterialsProject} and introduced by CDVAE~\cite{cdvae}. MP-20 contains 45,231 inorganic crystal structures with up to 20 atoms per unit cell, spanning 89 elements. Following CDVAE, we use the standard 60/20/20 split for training, validation, and testing. Since MP-20 consists mostly of experimentally known and globally stable materials, it provides a challenging benchmark for de novo crystal generation.
    \item \textbf{Alex-MP-20} is a large-scale dataset of inorganic crystal structures curated in MatterGen~\cite{mattergen} by combining data from the Alexandria database~\cite{alexandria} and the Materials Project~\cite{MaterialsProject}. The dataset includes 607,684 structures with at most 20 atoms per unit cell, ensuring compatibility with standard crystal generation benchmarks. Additional filtering is applied to retain thermodynamically stable or near-stable materials, typically defined as having energy above hull below 0.1 eV/atom, based on density functional theory (DFT) calculations. Materials containing radioactive atoms are removed.
\end{itemize}

\subsection{Evaluation Metric}
Our goal in de novo generation task it to generate valid, stable, unique and novel materials. These four basic metrics are defined as follows:
\begin{itemize}
    \item \textit{Validity} (V). We consider the validity of a crystal from both structure and composition~\cite{cdvae}. For structure, a valid crystal should have volume larger than 0.1, and the minimal distance among all atom pairs should larger than 0.5. For element composition, we check charge neutrality and electronegativity difference. If one crystal satisfies these two validity simultaneously, it is overall valid.
    \item \textit{Stability} (S). Stability measures whether a generated crystal is thermodynamically feasible. For each generated structure, we compute its energy above the convex hull, denoted as $\Delta E$, using a surrogate model (e.g., MLFF) or DFT when available. A structure is considered stable if $\Delta E<\epsilon$, where $\epsilon$ is a small threshold.
    \item \textit{Uniqueness} (U). Uniqueness measures the diversity of generated structures by removing duplicates within the generated set. Two generated structures are considered identical if they match under the same structural equivalence criterion.
    \item \textit{Novelty} (N). Novelty evaluates whether generated crystals are distinct from the reference dataset. A generated structure is considered non-novel if it matches any structure in the reference set under a structural equivalence criterion (e.g., using \texttt{StructureMatcher}). Otherwise, it is regarded as novel.
\end{itemize}
We discuss more on stability and novelty in this paper. To further evaluate generation quality, we further report compound metrics that measure the joint satisfaction of multiple criteria. Let $\{\mathbf{C}_{gen}\}_{i=1}^N$ denote $N$ generated crystals. For each crystal, we define indicator functions:
\begin{align}
S_i = \mathbb{I}(E_{\text{hull}}^{(i)} \leq \epsilon), \quad
N_i = \mathbb{I}(\mathbf{C}_i \notin \mathcal{D}_{\text{ref}}),  \quad
U_i = \mathbb{I}(\mathbf{C}_i \text{ is unique in } \mathcal{C}),  \quad
V_i = \mathbb{I}(\mathbf{C}_i \text{ is valid}).
\end{align}
Then, we define
\begin{itemize}
    \item \textit{Stable \& Unique \& Novel} (S.U.N) measures the fraction of generated crystals that are simultaneously stable, unique, and novel:
    \begin{equation}
        \texttt{S.U.N} = \frac{1}{N} \sum_{i=1}^N S_i \cdot U_i \cdot N_i.
    \end{equation}

    \item \textit{Valid \& Stable \& Unique \& Novel} (V.S.U.N) further incorporates validity, measuring the fraction of samples that satisfy all four criteria:
    \begin{equation}
        \texttt{V.S.U.N} = \frac{1}{N} \sum_{i=1}^N V_i \cdot S_i \cdot U_i \cdot N_i.
    \end{equation}
\end{itemize}
These compound metrics explicitly characterize the trade-off between stability and novelty, which is often overlooked when reporting marginal metrics independently.

\subsection{Density Functional Theory Settings}
\label{dft}
We use DFT settings from Materials Project \url{https://docs.materialsproject.org/methodology/materials-methodology/calculation-details/gga+u-calculations/parameters-and-convergence} for structure relaxation and energy computation. In particular, we do GGA and GGA+U calculations with \texttt{atomate2.vasp.flows.mp. MPGGADoubleRelaxStaticMaker}~\citep{ganose2025_atomate2}, which in turn relies on \texttt{pymatgen.io.vasp.sets.MPRelaxSet} and \texttt{pymatgen.io.vasp.sets.MPStaticSet}~\citep{ong2013python}. Computations themselves were done with VASP~\citep{kresse1996vasp} version 5.4.4. with the plane-wave basis set~\citep{kresse1996vasp}. The electron-ion interaction is described by the projector augmented wave (PAW) pseudo-potentials~\citep{kresse1999paw}. The exchange-correlation of valence electrons is treated with the Perdew-Burke-Ernzerhof (PBE) functional within the generalized gradient approximation (GGA)~\citep{perdew1996pbegga}. The raw total energies computed by DFT were corrected with \texttt{MaterialsProject2020Compatibility} before putting into the \texttt{PhaseDiagram} to obtain the DFT $E_\text{hull}$.

We do DFT relaxation firstly using the generated crystal structures. Then, for the crystals failed to DFT, we follow previous studies~\cite{flowmm,adit} and use MatterSim-v1-1M~\cite{mattersim} to do pre-relaxation, and then redo DFT.

\subsection{Distance based on Fingerprints}
\label{des_finger}
\subsubsection{CrystalNN and Magpie Fingerprints}
\textbf{CrystalNN fingerprints}~\cite{zimmermann2020local} are local structural descriptors implemented in matminer~\cite{ward2018matminer}, based on the neighbor-finding algorithm CrystalNN proposed in pymatgen~\cite{ong2013python}. For each atomic site, CrystalNN identifies neighboring atoms using a combination of distance-based criteria and Voronoi-like weighting, yielding a robust coordination environment even in distorted structures. Based on the identified neighbors, the fingerprint encodes features such as coordination number distributions, local bonding environments and weighted neighbor statistics. These site-level features are then aggregated (e.g., mean, variance) across all atoms in the unit cell to obtain a fixed-length vector representation for the entire crystal. CrystalNN fingerprints primarily capture local geometric and coordination information, making them effective for representing structural environments.

\textbf{Magpie fingerprints}~\cite{ward2016general} are composition-based descriptors introduced in matminer and originally proposed in the Magpie framework. Given only the chemical composition of a material, Magpie computes statistical summaries of elemental properties, including atomic number electronegativity, atomic radius, melting temperature, valence electron counts. For each property, Magpie derives aggregate statistics such as mean, variance, minimum / maximum and range. These statistics form a fixed-length feature vector describing the composition. Magpie fingerprints capture global compositional characteristics but do not encode explicit structural information.

\subsubsection{Definition of Distance}
To quantify the distances between $\{\mathbf{C}_{gen}\}$ and $\{\mathbf{C}_{gt}\}$ as in Section~\ref{tradeoff}, inspired by the \texttt{Precision} metric~\cite{cdvae}, we consider fingerprint based distance, and measure as follows:
\begin{itemize}
    \item For crystal $\mathbf{C}_i\in\{\mathbf{C}_{gen}\}$, we get its two vectors, i.e., $FP_i^{str}\in\mathbb{R}^{132}$ and $FP_i^{com}\in\mathbb{R}^{61}$. The former is CrystalNN fingerprint~\cite{zimmermann2020local} for structural information, and the latter is normalized Magpie fingerprint~\cite{ward2016general} describing elements composition.
    \item Compute the structure distance $\mathcal{D}_i^{str}= \min_{\mathbf{C}_j\in\{\mathbf{C}_{gt}\}}||FP_i^{str}-FP_j^{str}||^2$ and composition distance $\mathcal{D}_i^{com}=\min_{\mathbf{C}_k\in\{\mathbf{C}_{gt}\}}||FP_i^{com}-FP_k^{com}||^2$.
    \item Get the overall distance $\mathcal{D}_i=\mathcal{D}_i^{str}/\alpha+\mathcal{D}_i^{com}/\beta$, where $\alpha=\max_{\mathbf{C}_j\in\{\mathbf{C}_{gen}\}}\mathcal{D}_j^{str}$ and $\beta=\max_{\mathbf{C}_k\in\{\mathbf{C}_{gen}\}}\mathcal{D}_k^{com}$ for normalization.
\end{itemize}
The distance $\mathcal{D}_i$ jointly captures structural and compositional deviation from the training set.

\subsection{Operating Environment}
\label{Operating environment}
The environment where our code runs is shown as follows:
\begin{itemize}
    \item Operating system: Linux version 6.8.0-63-generic
    \item CPU information: AMD EPYC 9554 64-Core Processor
    \item GPU information: NVIDIA Corporation AD102GL [L40S]
\end{itemize}
\section{Related Work}
\label{related work}
\paragraph{Generation on Scientific Discovery}
Recent advances in deep generative models, particularly diffusion-based approaches, have significantly accelerated progress in scientific discovery, including molecule and material design. AlphaFold3~\cite{abramson2024accurate} leverages diffusion to enable accurate all-atom biomolecular complex generation. GeoLDM~\cite{xu2023geometric} demonstrates the effectiveness of diffusion models for scientific data by capturing 3D geometric structures and producing physically consistent molecular samples. Graph-GRPO~\cite{zhu2026graph} further advances molecule generation by aligning graph flow models with task-specific objectives through reinforcement learning~\cite{sui2026conversation}.

In the domain of crystal generation, CDVAE~\cite{cdvae}, FlowMM~\cite{flowmm}, and ADiT~\cite{adit} introduce variational, flow matching, and latent diffusion paradigms, respectively. DiffCSP~\cite{diffcsp} proposes CSPNet, which has become a widely adopted equivariant denoising backbone. Subsequent works, such as DiffCSP++~\cite{diffcsppp}, SGEquiDiff~\cite{sge}, and SymmCD~\cite{symmcd}, incorporate physical priors, including space group constraints and crystallographic symmetry, into the generation process. MatterGen~\cite{mattergen} explores conditional diffusion for inverse material design based on target properties and symmetry. More recently, several studies integrate large language models with crystal generation~\cite{CrystalLLM, Flowllm, crysllmgen}.

Overall, prior work primarily focuses on designing more advanced generative architectures and objectives. In contrast, our work shifts the focus toward improving the quality of generated samples via a screening-and-refinement pipeline. The substantial improvements achieved even with a simple base model highlight the general applicability of our approach across different generative frameworks.

\paragraph{Joint-Embedding Predictive Architecture}
Joint-Embedding Predictive Architecture (JEPA)~\cite{jepa} is a self-supervised learning paradigm that learns representations by predicting target embeddings from context embeddings in latent space, instead of reconstructing raw inputs. This design encourages the model to capture high-level semantic information while discarding irrelevant details. 

JEPA has been successfully extended to multiple domains. I-JEPA~\cite{ijepa} learns semantic image representations by predicting masked regions from partial context. V-JEPA~\cite{vjepa} scales this paradigm to video data through temporal prediction. VL-JEPA~\cite{vljepa} aligns visual and textual modalities in a shared embedding space, while LLM-JEPA~\cite{llmjepa} extends the framework to language and code modeling. LeJEPA~\cite{lejepa} further introduces regularization techniques to stabilize representation learning and prevent collapse. Additionally, LeWorldModel~\cite{Leworldmodel} connects JEPA with world modeling by predicting future latent states.

In contrast to these works, which primarily focus on representation learning, we explore the application of JEPA in scientific data, specifically for crystal modeling. Our work demonstrates that JEPA can serve as an effective surrogate for energy-aware comparison and play a crucial role in generative screening.

\paragraph{Transformer}
Transformers~\cite{vaswani2017attention} have become a general-purpose architecture for modeling structured data due to their ability to capture global dependencies via self-attention. Representative variants include LLaMA~\cite{Llama}, which introduces architectural improvements for large-scale autoregressive generation, and Vision Transformer (ViT)~\cite{vit}, which extends Transformers to image modeling and enables advances in multimodal learning. Transformers have also been adapted for generative modeling, such as Diffusion Transformers (DiT)~\cite{dit}, which provide a strong backbone for diffusion models. In graph domains~\cite{zhang2026toward}, Graph Transformer (GT)~\cite{gt} first introduces self-attention to graph data, while subsequent works address its limitations. For example, CoBFormer~\cite{xing2024less} firstly reveals the over-globalization issue and enhances local inductive biases with theoretical guarantees, and Specformer~\cite{bo2023specformer} makes the first attempt to learn eigenvalues interaction via Transformer, triggering the fusion of attention and spectral domain.

In this work, we adopt a standard Transformer architecture as the backbone. Exploring more advanced Transformer variants to further improve \M\ remains an important direction for future work.


\end{document}